%% file: main.tex
\pdfoutput=1
\RequirePackage{fix-cm}
\documentclass[]{template}

\input{macros}

\title{\gtwopname: Scaling Gameplay Experience for Visual Agents through Environment--Player Co-Development}
\renewcommand{\shorttitle}{\gplogo[1.05em]\hspace{0.35em}\gtwopname: Scaling Gameplay Experience for Visual Agents through Environment--Player Co-Development}

\author[]{%
Xirui Jiang$^{1*}$,\hspace{0.4em}
Yuxi Chen$^{1*}$,\hspace{0.4em}
Rui Yang$^{1*\dagger}$,\hspace{0.4em}
Feng Luo$^{2*}$,\hspace{0.4em}
Pengcheng Wang$^{1}$,\hspace{0.4em}
Bingxuan Li$^{1}$,\hspace{0.4em}
Zihao Zhao$^{3}$\\
Yushan Li$^{1}$,\hspace{0.4em}
Han Wang$^{1}$,\hspace{0.4em}
Manling Li$^{4}$,\hspace{0.4em}
Yang Zhang$^{5}$,\hspace{0.4em}
Shiyu Chang$^{2}$,\hspace{0.4em}
Tong Zhang$^{1}$,\hspace{0.4em}
Huan Zhang$^{1}$%
}

\affiliation{%
$^1$UIUC,\hspace{6pt}
$^2$UC Santa Barbara,\hspace{6pt}
$^3$Georgia Tech,\hspace{6pt}
$^4$Northwestern University,\hspace{6pt}
$^5$IBM
\\[5pt]
\linkicon{https://game2player.github.io}{\gplogo[1.15em]}{Website}\hspace{14pt}
\linkicon{https://github.com/game2player/Game2Player}{\includegraphics[height=1.05em]{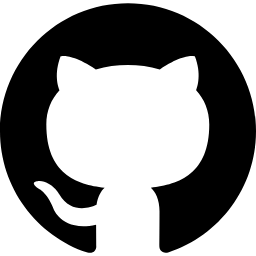}}{Code}\hspace{14pt}
\linkicon{https://huggingface.co/game2player}{\hficon{1.15em}}{Models}
}

\setfirstpagefooter{%
$^*$Equal contribution. $^{\dagger}$Project lead. Emails: \texttt{xiruij@illinois.edu}, \texttt{ry21@illinois.edu}%
}

\begin{document}

\setteaser{%
\centering
\begin{tikzpicture}
  \node[inner sep=0pt,outer sep=0pt,anchor=south west,opacity=0] (w) at (0,0) {\includegraphics[width=\linewidth]{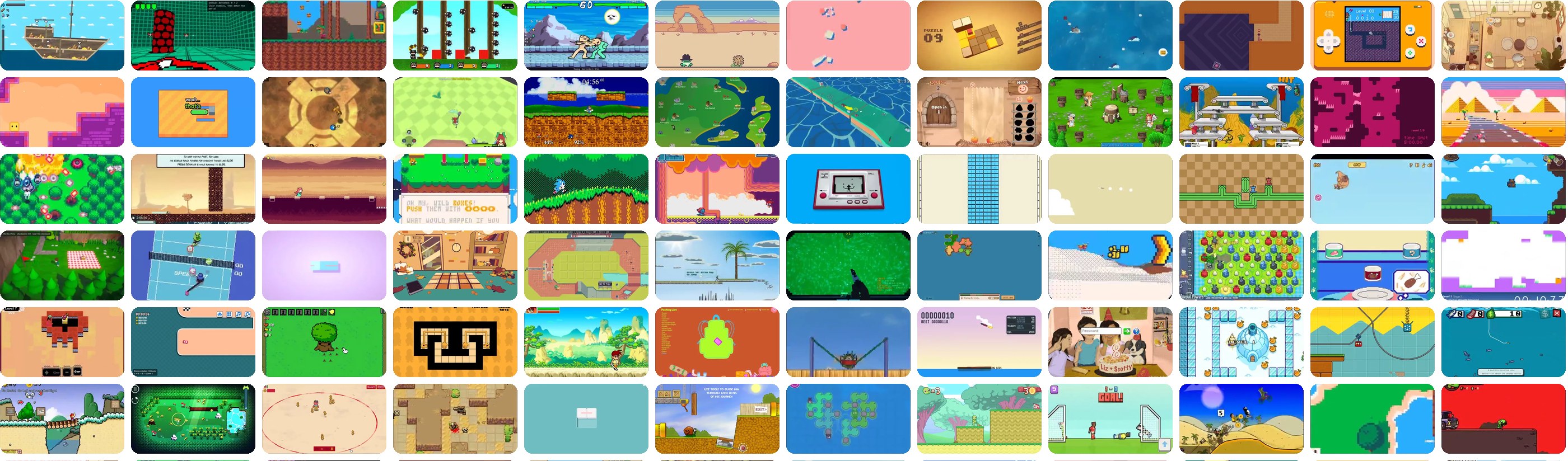}};
  \begin{scope}
    \clip[rounded corners=6pt] (w.south west) rectangle (w.north east);
    \node[inner sep=0pt,outer sep=0pt,anchor=south west] at (0,0) {\includegraphics[width=\linewidth]{figures/banner_wall.jpg}};
    \fill[white,opacity=0.42] (w.south west) rectangle (w.north east);
  \end{scope}
  \fill[white,opacity=0.9,rounded corners=8pt] ($(w.center)+(-6.75,-1.08)$) rectangle ($(w.center)+(6.75,1.08)$);
  \draw[g2pplayer!35,line width=0.7pt,rounded corners=8pt] ($(w.center)+(-6.75,-1.08)$) rectangle ($(w.center)+(6.75,1.08)$);
  \node[anchor=center,inner sep=0pt] at ($(w.center)+(0,0.30)$) {%
    \gplogo[1.0cm]\hspace{0.28cm}%
    {\fontsize{28}{28}\selectfont\sffamily\bfseries\gtwopname}};
  \node[anchor=center,inner sep=0pt,font=\normalsize\sffamily,text=g2pink!85] at ($(w.center)+(0,-0.62)$) {%
    \textcolor{g2pgame}{\bfseries 568} web games\qquad
    \textcolor{g2pgame}{\bfseries 3,436} verified tasks\qquad
    \textcolor{g2pplayer}{\bfseries 15.7K} trajectories\qquad
    \textcolor{g2pplayer}{\bfseries 307} hours of play};
\end{tikzpicture}\par
\vskip 0.5mm
\captionof{figure}{\sys turns \textbf{568} in-the-wild web games into verifiable environments with \textbf{3,436} state-verified tasks. The \cp and the \vp play them to collect \textbf{15.7K} goal-labelled trajectories.}
\label{fig:hero}
}
\paperabstract{\input{sections/0_abstract}}

\maketitle

\input{sections/1_intro}
\input{sections/2_related}

\input{sections/3_method}
\input{sections/4_experiments}
\input{sections/5_conclusion}

\bibliographystyle{plainnat}
\bibliography{main}

\ifincludesupp
\input{sections/X_suppl}
\fi

\end{document}

%% file: macros.tex
\usepackage[utf8]{inputenc}
\usepackage{amsmath,amssymb,amsfonts}
\usepackage{url}
\usepackage{nicefrac}
\usepackage{pifont}
\usepackage{array}
\usepackage{makecell}
\usepackage{tabularx}
\usepackage{colortbl}
\usepackage{needspace}
\usepackage{xspace}
\usepackage{wrapfig}
\usepackage{enumitem}
\usepackage{pgfplots}
\pgfplotsset{compat=1.17}
\usetikzlibrary{arrows.meta,positioning,calc,shapes.misc,fit,backgrounds,svg.path}
\usepgfplotslibrary{groupplots}
\usepackage[numbers,sort&compress]{natbib}
\IfFileExists{fontawesome5.sty}{\usepackage{fontawesome5}}{}

\definecolor{airforceblue}{rgb}{0.36, 0.54, 0.66}
\definecolor{bluegray}{rgb}{0.4, 0.6, 0.8}
\definecolor{bleudefrance}{rgb}{0.19, 0.55, 0.91}
\usepackage{hyperref}
\hypersetup{colorlinks,linkcolor={airforceblue},citecolor={bleudefrance},urlcolor={bluegray}}
\usepackage[noabbrev,nameinlink]{cleveref}
\crefname{section}{Sec.}{Secs.}
\Crefname{section}{Section}{Sections}
\crefname{figure}{Fig.}{Figs.}
\crefname{table}{Tab.}{Tabs.}
\crefname{equation}{Eq.}{Eqs.}
\crefname{appendix}{App.}{Apps.}

\definecolor{oursrow}{RGB}{226,239,252}
\definecolor{baserow}{RGB}{244,245,247}
\definecolor{codecol}{HTML}{2E7D5B}%
\definecolor{vlmcol}{HTML}{C0682B}%
\definecolor{basecol}{HTML}{8A8F98}%
\definecolor{g2pcol}{HTML}{177FCC}%
\definecolor{g2p9col}{HTML}{0B4F82}%

\definecolor{g2pgame}{HTML}{12A594}
\definecolor{g2pturn}{HTML}{E54D2E}
\definecolor{g2pplayer}{HTML}{5B5BD6}
\definecolor{g2pink}{HTML}{1F2328}
\makeatletter
\DeclareRobustCommand{\gpnamefont}{%
  \edef\gp@sf{\sfdefault}%
  \ifx\f@family\gp@sf\else
    \fontsize{\the\dimexpr\f@size pt*88/100\relax}{\f@baselineskip}\selectfont
  \fi
  \sffamily\bfseries}
\makeatother
\newcommand{\gtwopname}{{\gpnamefont\textcolor{g2pgame}{Game}\textcolor{g2pturn}{2}\textcolor{g2pplayer}{Player}}}
\newcommand{\pbname}{{\gpnamefont\textcolor{g2pplayer}{Player}\textcolor{g2pink}{Bench}}}
\newcommand{\sys}{\texorpdfstring{\gtwopname}{Game2Player}\xspace}
\newcommand{\bench}{\texorpdfstring{\pbname}{PlayerBench}\xspace}
\newcommand{\gw}{GameWorld\xspace}
\newcommand{\cp}{Coding Player\xspace}
\newcommand{\vp}{VLM Player\xspace}
\newcommand{\gfour}{Game2Player-4B\xspace}%
\newcommand{\gnine}{Game2Player-9B\xspace}
\newcommand{\gapi}{state API\xspace}
\newcommand{\phase}[1]{\textsc{#1}}

\newcommand{\cmark}{\textcolor{green!55!black}{\ding{51}}}
\newcommand{\xmark}{\textcolor{red!70!black}{\ding{55}}}
\newcommand{\pmark}{\textcolor{orange!80!black}{$\circ$}}

\newcommand{\gplogo}[1][1em]{\begingroup\pgfmathsetmacro{\gps}{(#1)/24}%
\tikz[baseline=0.16*#1,x=1pt,y=1pt]{\begin{scope}[scale=\gps,yscale=-1,shift={(0,-24)}]
\fill[g2pplayer,rounded corners=6*\gps pt] (0,0) rectangle (24,24);
\draw[white,line width=1.5*\gps pt,line join=round] svg{M7.5 7.5h9a3 3 0 0 1 2.9 2.3l1 5a2 2 0 0 1-3.3 1.9L14.5 15h-5l-2.6 1.7a2 2 0 0 1-3.3-1.9l1-5a3 3 0 0 1 2.9-2.3Z};
\draw[white,line width=1.5*\gps pt,line cap=round] svg{M8 10v4m-2-2h4};
\filldraw[fill=g2pgame,draw=white,line width=0.6*\gps pt,line join=round] svg{M14 10l4 2.5-4 2.5z};
\end{scope}}\endgroup}

\newcommand{\hficon}[1]{\includegraphics[height=#1]{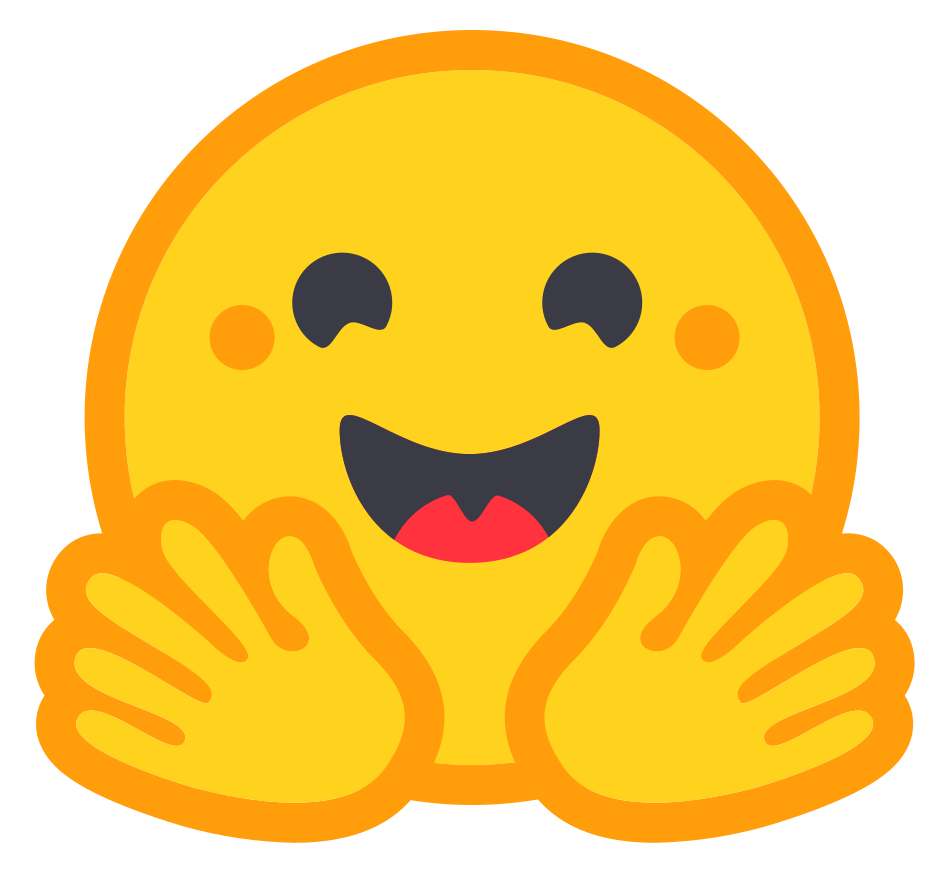}}

\newcommand{\linkicon}[3]{\href{#1}{\raisebox{-0.22em}{#2}\hspace{0.3em}{\sffamily\color{g2pplayer}#3}}}

\colorlet{tblink}{blue}
\colorlet{tblaccent}{blue}
\definecolor{tblmuted}{HTML}{65727D}
\colorlet{tblrule}{blue!35!white}
\colorlet{tblheadbg}{blue!6!white}
\colorlet{tblgroupbg}{blue!3!white}
\colorlet{tbloursbg}{blue!10!white}
\colorlet{tblbasebg}{blue!2!white}
\newcommand{\compacttable}[1][7.8]{%
  \sffamily\fontsize{#1}{\dimexpr #1pt + 1.6pt\relax}\selectfont
  \setlength{\parskip}{0pt}%
  \setlength{\parindent}{0pt}%
  \setlength{\tabcolsep}{3.5pt}%
  \renewcommand{\arraystretch}{1.04}%
  \setlength{\extrarowheight}{0pt}%
  \setlength{\aboverulesep}{1.6pt}%
  \setlength{\belowrulesep}{1.8pt}%
  \setlength{\heavyrulewidth}{0.65pt}%
  \setlength{\lightrulewidth}{0.32pt}%
  \setlength{\cmidrulewidth}{0.32pt}%
  \arrayrulecolor{tblrule}%
  \renewcommand{\cmark}{\textcolor{tblaccent}{\ding{51}}}%
  \captionsetup{font={footnotesize,rm},singlelinecheck=false,justification=raggedright,position=bottom,skip=4pt}%
}
\newcommand{\tblhead}[1]{{\leavevmode\sffamily\bfseries\color{tblink}#1}}
\newcommand{\tblbold}[1]{{\leavevmode\bfseries #1}}

\newcommand{\tblsection}[2]{\multicolumn{#1}{@{}l}{\cellcolor{tblgroupbg}\tblhead{#2}}\\}

\titleformat{\paragraph}[runin]{\bffont\bfseries\sffamily}{}{0pt}{}
\titlespacing*{\paragraph}{0pt}{0.9ex plus 0.3ex minus 0.2ex}{0.5em}

\usepackage{fancyvrb}
\usepackage{multicol}
\definecolor{rolevlm}{HTML}{D0662A}\definecolor{rolevlmbg}{HTML}{FEF6F0}
\definecolor{rolebuilder}{HTML}{2F6DB5}\definecolor{rolebuilderbg}{HTML}{F2F7FC}
\definecolor{rolecoder}{HTML}{2E8B57}\definecolor{rolecoderbg}{HTML}{F1F9F4}
\definecolor{rolescout}{HTML}{8A6D1A}\definecolor{rolescoutbg}{HTML}{FCF8EC}
\newcommand{\promptrole}{black}
\newcommand{\promptcmds}{%
  \def\hd##1{{\sffamily\bfseries\textcolor{\promptrole}{##1}}}%
  \def\ph##1{\textcolor{black!55}{\textit{##1}}}%
  \def\om##1{\textcolor{black!40}{##1}}%
  \def\bf##1{{\sffamily\bfseries ##1}}}
\newcommand{\promptfile}[1]{\VerbatimInput[fontsize=\fontsize{6.5}{7.4}\selectfont,commandchars=\|\@\!,formatcom=\promptcmds]{#1}}
\newtcolorbox{promptbox}[2]{enhanced,colframe=#1,colback=#1bg,colbacktitle=#1,coltitle=white,
  boxrule=0.6pt,arc=2.5pt,left=4pt,right=4pt,top=2.5pt,bottom=2.5pt,toptitle=1pt,bottomtitle=1pt,
  fonttitle=\sffamily\bfseries\scriptsize,title={#2},before upper={\renewcommand{\promptrole}{#1}}}
\newif\ifincludesupp \includesupptrue
\newcommand{\colfigwidth}{0.62\linewidth}

%% file: sections/0_abstract.tex
Games are a rich testbed for visual agents, demanding perception, long-horizon planning, and precise control. However, collecting human gameplay is costly, and learning from gameplay videos requires inferring missing actions. Existing datasets also cover a limited range of games, often without explicit tasks or verifiable outcomes. To address these gaps, we introduce \textbf{\sys}, a framework that \emph{co-develops} game environments and players to collect diverse and verifiable gameplay experience at scale. In the \sys loop, a Builder agent adapts existing web games, exposes their internal state, and defines tasks with state-based progress measures and success checks. A \cp and a \vp then attempt the tasks and report problems to the Builder for repair, so that play serves as both environment validation and training data collection. With \sys, we build 568 verified web games with 3,436 tasks, collect 15.7K trajectories (307 hours of gameplay), and hold out 50 games as the benchmark \textbf{\bench}. To evaluate the collected experience as training data, we train VLM agents with \emph{progress-weighted SFT}, assigning greater weight to steps that advance task progress. \gnine improves progress from 14.3\% to 31.2\% on \bench and from 30.0\% to 53.1\% on the external GameWorld benchmark. More broadly, our work suggests a path for visual agents to scale their gameplay experience by building, verifying, and learning from existing game worlds.

%% file: sections/1_intro.tex
\section{Introduction}
\label{sec:intro}

Visual agents powered by vision-language models (VLMs) have made rapid progress on computer-use and embodied tasks~\cite{qin2025ui,yang2025embodiedbench,li2026memcompress,wang2025ui,szot2025multimodal}. Video games pose a broader challenge for visual intelligence~\cite{arxiv240303186,arxiv230516291,arxiv240410179,arxiv251023691,arxiv251108892,arxiv260102427}: each game brings its own visuals, rules and controls, which an agent must infer from the screen while reasoning over a changing scene and acting with precise timing. This diversity makes gameplay experience a valuable source of training data for general-purpose visual agents. We therefore aim to collect gameplay data at scale across diverse games for visual agent training.

A central bottleneck is obtaining such gameplay experience at scale: learning requires trajectories that pair actions with explicit goals and reliable signals of success. Human gameplay recordings~\cite{guss2019minerl,zhang2019atari,kurin2017atari,pearce2021counter,he2025plaicraft} are expensive to collect, and online gameplay videos lack actions, which have to be recovered with inverse-dynamics or latent-action models~\cite{baker2022video,schmidt2023learning,choi2025d2e,yue2025pixels}; in both cases, goals and outcomes are annotated after the fact~\cite{simateam2025sima,arxiv260925001,yue2026scaling,arxiv240410179}. Interactive game environments provide all three natively~\cite{arxiv241113543,arxiv250518134,arxiv250515146,arxiv250603610,arxiv260407429,arxiv260306656,arxiv250901052,arxiv260609826}, but each new game demands game-specific engineering, which limits their coverage.

Coding agents could automate both the environment building and gameplay needed for data collection.
They can synthesize verifiable environments from code repositories and websites~\cite{yang2025swesmith,chae2026safeandscala,wang2026cuagym,wu2026autowebworld}, write games from scratch~\cite{arxiv260217594,arxiv260922308}, and develop executable policies for playing games~\cite{arxiv230516291}.
Adapting existing web games for such data collection requires locating their internal state in engine-specific code and using it to define task initialization, progress measures and success checks.
However, the resulting environment may pass the coding agent's own tests even when a task is impossible to complete, its progress measure fails to update, or its success check triggers too early.
\emph{Such defects may only become apparent when an agent actually plays the task.}
This motivates \emph{environment--player co-development}: environments and the players that test them are developed together, each improving the other through play.

Building on this principle, we present \textbf{\sys}, a framework in which a \textbf{Builder} agent and two players, a \textbf{\vp} and a \textbf{\cp}, co-develop verifiable environments from in-the-wild web games (\cref{fig:overview}).
The Builder sets up each game in a sandboxed, reproducible runtime, exposes its internal state, and defines tasks whose progress and success are read from that state, guided by engine-specific skills.
The players then attempt every task: the \vp plays from screenshots, pausing the game while it reasons, whereas the \cp writes a Python policy over player-visible state, runs it in real time and refines the code from the recorded outcomes.
When a failure traces back to the environment, the players report it, either as a defect, such as a key with no effect or progress that never moves, or as a request for on-screen information that the policy cannot read.
The Builder checks each report against the recorded episode before repairing the environment, and accepts an environment once the players' episodes confirm its own evaluation (\cref{sec:codev}).

Once an environment is accepted, the players turn from testing it to collecting experience in it, and every recorded episode is paired with its task instruction and a dense, state-based progress signal.
The two players are complementary teachers, offering efficient demonstration collection and visually grounded action explanations, respectively.
A code policy can be rerun without any model calls, supplying high-quality demonstrations quickly and at scale.
The \vp acts on the same screenshots that the trained agent will see and explains each action in a short note, so its episodes show how to approach a game even when they fall short of the goal.
We distill both sources into one pixel-based agent with \emph{progress-weighted SFT}, which up-weights actions that advance task progress.

Using \sys, we have built 568 verified environments with 3,436 tasks across more than 25 genres and seven engines.
As verifiable game benchmarks remain limited in both number and diversity, we also use it to construct \textbf{\bench}, 50 held-out games with 287 tasks for evaluating game agents on unseen games.
In a controlled rebuild of 20 games, player reports raise the share of correctly implemented tasks from 65\% to 72\% and the tasks solved by code policies from 34 to 46.
The code policies written in the loop play in real time without model calls and record 22 times as many steps per hour as a \vp.
Training on 15.7K trajectories more than doubles the \bench progress of Qwen3.5-9B, from 14.3\% to 31.2\%, and the gains transfer to the external benchmark \gw~\cite{arxiv260407429}, where \gnine reaches 53.1\%, higher than the published results of several proprietary models.
Beyond these results, our ablations examine how progress-weighted SFT, the data from each player, and thinking at inference time contribute to the gains.

In summary, our contributions are:
\begin{itemize}[leftmargin=1.2em,itemsep=2pt,topsep=3pt]
\item \textbf{Environment--player co-development}: \sys, a scalable framework where a Builder and two players use gameplay feedback to build and validate verifiable environments;
\item \textbf{\bench}: 50 held-out games and 287 tasks, more challenging than \gw, where agents of \cref{tab:players} score on average 12.8 and 7.6 points lower progress and success;
\item \textbf{Progress-weighted distillation}: training on 15.7K trajectories from both players more than doubles \mbox{Qwen3.5-9B}'s progress on \bench, with gains transferring to \gw.
\end{itemize}

%% file: sections/2_related.tex
\section{Related Work}
\label{sec:related}

\noindent\textbf{Interactive game benchmarks.}
Existing game benchmarks span text and grid games~\cite{arxiv231001557,arxiv241113543,arxiv260324621}, puzzles and adventures~\cite{arxiv240713729,arxiv250901052}, browser games~\cite{arxiv250218356,arxiv260407429,arxiv250515146}, and commercial or 3D environments~\cite{arxiv250518134,arxiv240303186,arxiv260609826}.
\gw~\cite{arxiv260407429} provides state-based verification for 34 browser games; AI GameStore~\cite{arxiv260217594} generates 100 games with human involvement; and PlaySuite~\cite{arxiv261007127} evaluates thousands of open-source games with a video-language judge.
\sys automates the adaptation of existing games into state-verified environments for both training and evaluation.

\noindent\textbf{Offline game benchmarks and gameplay data.}
Question-answering benchmarks on gameplay video~\cite{mun2016marioqa,taesiri2023glitchbench,cao2024physgame,taesiri2025videogameqa,wang2026gameplayqa,arxiv260925001} test perception without play, and strong offline scores do not imply strong play~\cite{arxiv250602387,arxiv260925001}.
Gameplay corpora with recorded or inferred actions train generalist game agents~\cite{baker2022video,arxiv260102427,choi2025d2e,yue2025pixels,li2025jarvis,arxiv251023691}, and some add language annotations afterwards~\cite{arxiv240410179,simateam2025sima,arxiv260925001}.
Synthetic games with verifiable rewards improve VLM reasoning~\cite{arxiv250513886,arxiv250608011,arxiv250513426}.
Our trajectories are collected \emph{against} a task, so every step carries its instruction, its state-grounded progress and its outcome.
\Cref{tab:compare} compares \sys with game benchmarks and gameplay datasets.

\noindent\textbf{Environment synthesis for agents.}
Automated pipelines construct verifiable environments from code repositories~\cite{yang2025swesmith,jain2025r2egym}, terminal tasks~\cite{arxiv260116443,arxiv260710891}, websites and applications~\cite{chae2026safeandscala,wang2026cuagym,wu2026autowebworld}, or recreate games from observation~\cite{arxiv260922308}.
Open-ended learning further co-evolves environments and agents~\cite{arxiv190101753,arxiv201202096,faldor2025omniepic,zala2024envgen}.
In \sys, the feedback that drives repairs comes from two players with different views who play every task of an existing game, and the Builder checks their failures against the recordings before repairing the environment.
We apply co-development to refine the environments while developing two complementary players for scalable data collection.

\noindent\textbf{Training visual agents.}
Visual agents learn from trajectories through supervised fine-tuning and reinforcement learning across embodied, computer-use and game environments~\cite{chen2025era,xu2024aguvis,wang2025opencua,yang2026gui,chen2026recap,yang2026openwebrl,arxiv260102427}.
Concurrent work scales gameplay training: PLAYER-1~\cite{ouyang2026player1} uses demonstrations from 172 games followed by RL in 16 synthetic environments, while GameScaling~\cite{zhang2026gamescaling} continues pre-training on up to 1,600 hours of gameplay.
We instead collect complementary demonstrations from code policies and a visual player in verified environments and distill them through progress-weighted SFT, achieving higher progress and success rates on both \bench and \gw.

\noindent\textbf{Code policies, privileged teachers and reasoning.}
Code can serve directly as a game agent's policy~\cite{arxiv230516291,arxiv250313356}.
Distilling a teacher with privileged information into a student that sees less is a long-standing recipe in embodied learning~\cite{weihs2020bridging,choudhury2024better,penaloza2026privileged}.
We use state-reading code policies as teachers for pixel-based students and weight their supervision by measured task progress, related in spirit to step-level filtering of agent trajectories~\cite{he2025webstar}.
Our weights come from progress events measured in the game state, and they set how often each step is sampled.
Reasoning in agent trajectories is either produced by the agent itself~\cite{yao2022react,qin2025ui,arxiv251023691}, written afterwards~\cite{yang2024react,xu2024aguvis,wang2025opencua}, or grounded in privileged state~\cite{hu2023thought,zawalski2024robotic,choudhury2024better}.
\sys follows the first route, retaining the \vp's own notes and native thinking during play as supervision, whereas code policy demonstrations carry neither kind of text and supervise actions only.

\input{tables/tab_compare}

%% file: tables/tab_compare.tex
\begin{table}[t]
\centering\compacttable[7.2]
\begin{tabularx}{\linewidth}{@{}>{\raggedright\arraybackslash}p{0.245\linewidth}r >{\raggedright\arraybackslash}p{0.16\linewidth}*{4}{>{\centering\arraybackslash}X}@{}}
\toprule
\rowcolor{tblheadbg}\tblhead{Resource} & \tblhead{\# Games} & \tblhead{Actions} & \tblhead{Goal} & \tblhead{State} & \tblhead{Prog.} & \tblhead{Autom.} \\
\midrule
VideoGameBench~\cite{arxiv250518134} & 10 & -- & -- & \xmark & \cmark & \xmark \\
LMGame-Bench~\cite{arxiv250515146} & 6 & -- & -- & \cmark & \pmark & \xmark \\
\gw~\cite{arxiv260407429} & 34 & -- & -- & \cmark & \cmark & \xmark \\
PlaySuite~\cite{arxiv261007127} & 5,734 & -- & -- & \xmark & \cmark & \pmark \\
\addlinespace[2pt]
VPT~\cite{baker2022video} & 1 & Human & \xmark & \xmark & \xmark & -- \\
NitroGen~\cite{arxiv260102427} & $>$1K & Human & \xmark & \xmark & \xmark & -- \\
PLAYER-1~\cite{ouyang2026player1} & 172 & Human & \cmark & \pmark & \xmark & \xmark \\
GameScaling~\cite{zhang2026gamescaling} & 80 & Human & \xmark & \xmark & \xmark & -- \\
\midrule
\rowcolor{tbloursbg}\sys (ours) & 568 & Code+VLM & \cmark & \cmark & \cmark & \cmark \\
\bottomrule
\end{tabularx}
\par
\caption{\textbf{Comparison with game resources.} Goal: demonstrations carry task instructions. State: scoring reads internal game state. Prog.: intermediate progress is scored. Autom.: environments or evaluators are built automatically.\cmark{} yes, \pmark{} partial, \xmark{} no, -- not applicable.}
\label{tab:compare}
\end{table}

%% file: sections/3_method.tex
\section{\sys}
\label{sec:method}
\label{sec:pipeline}

\sys turns games that are open to download from the web into verifiable environments and records gameplay in them for training (\cref{fig:overview}).
The Scout finds games that match a request, either a description or a game name, and that one person can play offline in a browser (\cref{supp:construction}).
The Builder, a coding agent, turns each game into an environment, and a \vp and a \cp play every task.
\Cref{sec:env} defines environments, tasks and the two players.
\Cref{sec:codev} describes how the Builder and the players co-develop each environment, \cref{sec:experience} how both players record training data in accepted environments, and \cref{sec:distill} how progress-weighted SFT trains a visual agent on them.

\input{figures/float_overview}

\subsection{Environments, tasks and players}
\label{sec:env}
\label{sec:formulation}

We model a game as a partially observed process with hidden state $s_t$, keyboard and mouse actions $a_t$ and screenshots $o_t$.
A \emph{verifiable environment} wraps the game so that it runs offline from a fixed seed and pauses while a model chooses an action.
Through one interface, the \emph{\gapi}, it exposes two views of $s_t$.
The \emph{state probe} $\phi(s_t)$ is a small set of named variables read from the running game, such as the player's health, and only scoring reads it.
The \emph{player observation} $\psi(s_t)$, which code policies read, holds named fields, such as the positions of visible objects, limited to what a person could see or infer from the screen, with no hidden information such as face-down cards.

A task is a tuple
\begin{equation}
\begin{gathered}
\tau=(\iota,\,c,\,p,\,g),\\
p_t=p\bigl(\phi(s_t)\bigr)\in[0,1],\qquad g_t=g\bigl(\phi(s_t)\bigr)\in\{0,1\},
\end{gathered}
\label{eq:task}
\end{equation}
where $\iota$ is a natural-language instruction, $c$ a start routine that puts the game into the situation where the task begins, $p$ a progress function and $g$ a success predicate, with \mbox{$g_t=1\Rightarrow p_t=1$}.
Both $p$ and $g$ read only the state probe, so scoring needs no human or model judge.
An \emph{episode} is one attempt at a task.
It begins with the start routine, lasts at most $T$ steps, each one observation followed by one action, and scores the best progress $\bar p=\max_t p_t$ that it reaches.
Progress rises with the milestones of ordinary play, so an episode that falls short of success still earns partial credit.
In the task of \cref{fig:overview}, for example, the state probe reads the delivered iron (\texttt{resources.iron}), progress grows with each delivery, and success requires buying Drill 1.

The two players act on different views of the same game (\cref{fig:data}).
A \emph{\vp} is a vision-language model that sees only the screenshots $o_t$.
Every \vp plays through one \emph{harness}.
At each step the harness pauses the game and gives the model the task instruction, the current screenshot, its recent steps and the \emph{control guide}, the Builder's account of what each key and mouse action does.
The model returns one action with a \emph{note}, one sentence on why it chose the action.
The \cp is a coding agent that writes one \emph{code policy} per game, a Python program that reads only the player observation $\psi(s_t)$ and plays each of the game's tasks in real time without calling a model.
Either player's episode is recorded as a \emph{trajectory}
\begin{equation}
\xi=\bigl(\tau,\;\{(o_t,\,a_t,\,z_t,\,p_t)\}_{t=1}^{T}\bigr),
\label{eq:traj}
\end{equation}
where $z_t=(n_t,r_t)$ holds a \vp's note $n_t$ and, if any, its thinking $r_t$.

\input{figures/float_data}

\subsection{Environment--player co-development}
\label{sec:codev}
\label{sec:builder}
\label{sec:repair}

When the Builder tests its own environment, it uses the inputs it designed and reads the variables it exposed, so it can miss problems that only a player would meet.
A player who knows only the control guide and the screen may find a control that the guide does not explain or a quantity that a policy needs but cannot read.
Co-development therefore lets \emph{both players test every environment through the interfaces that later produce training data and evaluation scores}, and report what they find to the Builder.
The loop repeats the four stages of \cref{fig:overview} until the environment is accepted.

\noindent\textbf{Build.}
The Builder works in one persistent session with the game, its source and a library of \emph{engine skills}, instructions for hooking into engines such as Unity WebGL, Flash, Godot and PICO-8.
It first proposes a task plan backed by evidence from the source or from play.
The plan is then frozen: a task may be removed only if the Builder cannot implement it, and \emph{a task that proves hard can be neither dropped nor made easier}.
The Builder packages the game for offline play and writes the state probe, the control guide and the start routine, progress function and success predicate of every task.
Finally, it evaluates the environment without editing it.

\noindent\textbf{Play.}
The \vp plays every task through the harness with a fixed model.
The \cp runs the game's code policy on every task and revises it until the episodes succeed or their failures point to the environment.

\noindent\textbf{Report.}
A player reports each failure that it traces to the environment and cites the recorded episodes.
Either player can file a \emph{defect report}, which describes something that does not work, such as a menu that hides part of the game, a key with no effect, progress that never moves or success that fires too early.
Only the \cp files \emph{observation requests}, which name a quantity that a code policy needs but cannot read from the player observation, with its meaning and where it appears on the screen; a \vp reads the screen directly and needs none.
\emph{No report may ask for hidden information, a shortcut to success or an easier task.}

\noindent\textbf{Verify, then repair.}
Each report reopens the Builder.
A player can mistake its own error for a defect, so the Builder first replays the recorded episode and \emph{checks the claim against the video and the state trace}.
If the checks contradict the claim, the Builder marks it disproved and cites the evidence.
Otherwise it repairs the environment: a defect report leads to changes in the game code, the state probe or the control guide, and an observation request adds fields to the player observation.
The players then play again.

\noindent\textbf{Acceptance.}
The environment and the code policy are accepted separately.
An environment is accepted when \emph{every player report has been fixed or disproved} and the environment passes a scripted gate that re-checks it for obvious errors and shortcuts.
A code policy is accepted when it completes a sufficient share of the game's tasks, with at least two successful episodes for every task it claims (\cref{supp:construction}).

\input{figures/float_collection}

\noindent\textbf{The collection.}
Running the pipeline on games from itch.io, Newgrounds and other portals produced 568 accepted environments with 3,436 tasks, after the Scout skipped games that cannot be played offline by one player and the Builder removed tasks that it could not implement.
They span more than 25 genres and seven major engines (\cref{fig:collection}).
We hold out 50 of these games as \bench, our test set, and use the other 518 for training.

\subsection{Gameplay data from two players}
\label{sec:experience}

Both players record training gameplay in the accepted environments, and their data are complementary.
A code policy calls no model while it plays, so it can be rerun from new seeds at little cost, whereas \vp episodes keep the model's notes and thinking and also cover tasks that no code policy completes.
To train on both, we bring them into one format.
\vp episodes already follow the harness, one action per step.
A code policy acts continuously, so we cut its recording into harness steps (\cref{fig:data}, middle): the recorder logs every key press and release with its time, and we convert this log into harness actions.
In total, the two players recorded 15.7K trajectories with 307 hours of gameplay, with more details in \cref{tab:data-sources}.
Of the resulting 1.56M training examples, code policies provide 87.7\% and \vp{}s 12.3\%.

\subsection{Progress-weighted SFT}
\label{sec:distill}

\begin{figure}[t]
\centering
\includegraphics[width=\linewidth]{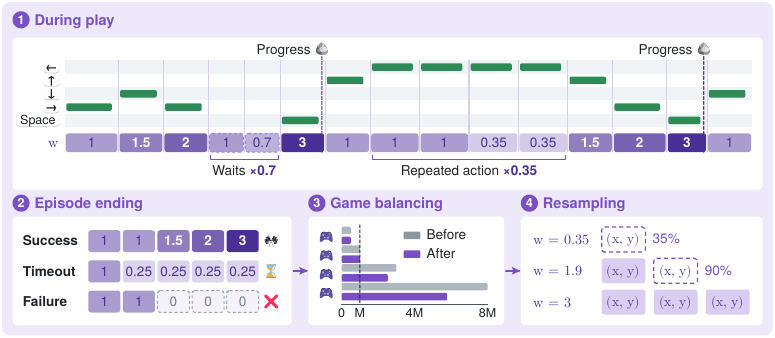}
\caption{\textbf{Progress-weighted sampling.} Actions receive weights from progress and repetition during play (1), followed by terminal outcomes (2). Paired bars compare the action budgets before and after game balancing (3), in units of the median game size $M$. Stochastic rounding converts the weights into integer numbers of context--target pairs $(x,y)$ (4). Solid cards are guaranteed copies, and dashed cards are drawn with the displayed probabilities. The examples are schematic and use the factors in \cref{tab:weighting}.}
\label{fig:weights}
\end{figure}

A trajectory spends long stretches waiting, wandering or recovering from mistakes, while only a few actions, such as a well-timed jump or the turn into a portal, move the task forward.
The progress $p_t$ recorded in every trajectory (\cref{eq:traj}) tells these actions apart, and \emph{progress-weighted SFT} uses it to weight the training examples.
Each example pairs the harness context $x_t$ at step $t$, the same input that the trained model receives when it plays, with a target $y_t$: the action $a_t$ for a code policy step, and the action together with the note and thinking, when present, for a \vp step.
Its weight is
\begin{equation}
w_t\;\propto\;\rho_t\cdot\max(1,\,c_t),
\label{eq:weight}
\end{equation}
where the credit $c_t$ raises the weight of the last few actions before each rise in progress or success, since these actions most likely caused it, and the penalty $\rho_t$ removes or lowers the weight of the final actions of failed and timed-out episodes and of repeated actions that do not improve progress.
The weights of games with many examples are also scaled down, so that a few long games do not dominate.
\Cref{fig:weights} illustrates these rules, and \cref{tab:weighting} gives their exact values and time windows.
Instead of weighting the loss, we round each weight stochastically to an integer multiplicity $m_t$ with $\mathbb{E}[m_t]=w_t$ and resample the examples accordingly, so training minimizes
\begin{equation}
\mathcal{L}(\theta)=-\sum_t m_t\log\pi_\theta(y_t\mid x_t).
\label{eq:loss}
\end{equation}
From the 2.05M eligible steps left after segmentation and filtering, this resampling draws 1.56M training examples.

%% file: figures/float_overview.tex
\begin{figure}[t]
\centering
\includegraphics[width=\linewidth]{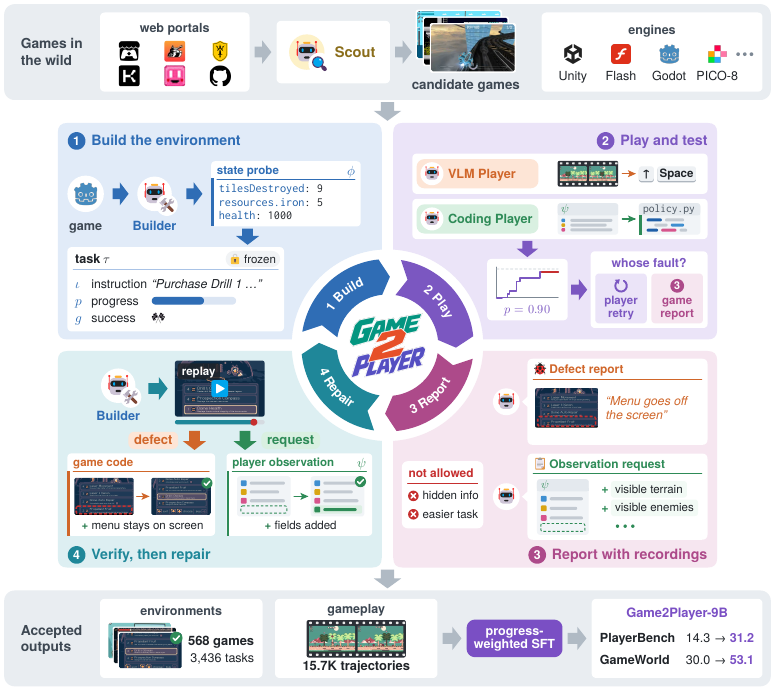}
\caption{\textbf{Overview of \sys.}
The Scout selects games from web portals (\textit{top}).
In the co-development loop (\textit{middle}), the Builder turns each game into an environment with state-verified tasks, the \vp and the \cp play every task and report faults of the game, and the Builder checks each report against the recording before repairing it.
Both players then record gameplay in the accepted environments to train \sys with progress-weighted SFT (\textit{bottom}).}
\label{fig:overview}
\end{figure}

%% file: figures/float_data.tex
\begin{figure*}[t]
\centering
\includegraphics[width=\linewidth]{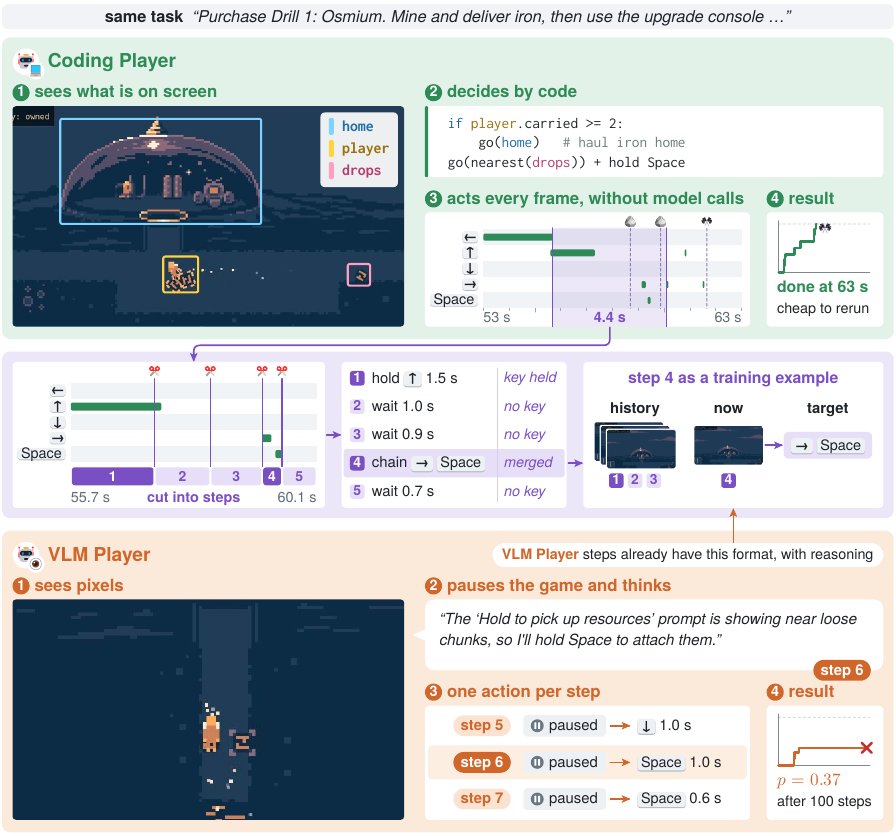}
\caption{\textbf{Two players on the same task and one training format.}
The code policy reads the player observation $\psi$ and sends keys at every frame without calling a model (\textit{top}).
The \vp sees only the screenshot and pauses the game to choose each action (\textit{bottom}).
The code policy's keys are cut into steps in the \vp's format, so both players yield the same kind of training example (\textit{middle}).}
\label{fig:data}
\end{figure*}

%% file: figures/float_collection.tex
\begin{figure}[t]
\centering
\includegraphics[width=\linewidth]{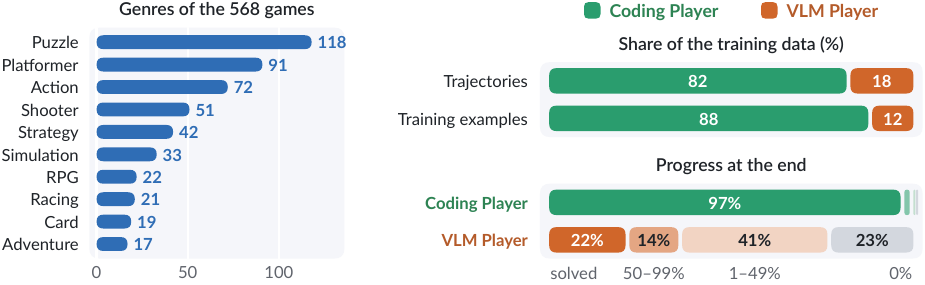}
\caption{\textbf{The \sys collection and its training data.} The 568 games cover diverse genres, of which the ten most common are shown (\textit{left}). Share of each player in the training data, and the best progress of its training trajectories (\textit{right}). \Cref{fig:collection-full} compares all genres and engines with \mbox{\bench}.}
\label{fig:collection}
\end{figure}

%% file: sections/4_experiments.tex
\section{Experiments}
\label{sec:experiments}

\subsection{Benchmarks and setup}
\label{sec:exp-setup}

\noindent\textbf{Benchmarks.}
\label{sec:collection}
We split the \sys collection by game and hold out 50 games with all 287 of their tasks as \bench, so no trajectory from a \bench game is used for training.
\bench covers seven engines and four difficulty tiers (\cref{supp:split}).
We also evaluate on \gw~\cite{arxiv260407429}, an independently built benchmark of 34 browser games and 170 tasks with state-based scoring and no overlap with our collection.
We evaluate \gw with our harness rather than its generalist one; neither is consistently better (\cref{tab:harness-comparison}).
On both, we report \emph{Progress}, the mean episode score $\bar p$ over equally weighted games, and \emph{Success}, the share of tasks completed.

\noindent\textbf{Protocol.}
A \vp attempts each task once from the same seed, playing through the harness at temperature 1 for at most 100 steps while the game pauses for each decision.
Our trained models and the open-weight baselines see 50 steps of text history and six earlier screenshots besides the current one, while API models receive their full text history.
Each step may use at most 512 output tokens with thinking off and 32,768 with thinking on.

\noindent\textbf{Training.}
We fine-tune Qwen3.5-4B and Qwen3.5-9B~\cite{qwen35} with progress-weighted SFT on our gameplay data, keeping the vision encoder frozen, which gives \mbox{\gfour} and \mbox{\gnine}.
Both sizes share one recipe: one epoch of 9,237 steps at learning rate $10^{-5}$ with cosine decay.
More details are given in \cref{supp:training}.

\subsection{Co-development builds correct and solvable environments}
\label{sec:exp-env}

\input{figures/float_loop}
\input{tables/tab_codev_repairs_main}

\noindent\textbf{Setup.}
We measure what the co-development loop adds by building the same 20 games with 145 frozen tasks in three settings (\cref{supp:loop}).
The \emph{full loop} runs \sys with a GPT-6.1 Terra Builder and GPT-6.1 Sol players, and reports return to the Builder for up to two rounds of repair.
The \emph{no-reports} setting shares the build and play of the full loop up to the first report, after which the environment stays as delivered and the \cp continues with the same budget.
The \emph{generic-agent} setting gives the same games and task requests to a generic coding agent with the Builder's model but without our skills and tools, and a second generic agent writes the code policy.
An auditor model checks every task against the original game and the frozen task definition, and we verify its verdicts by hand.
A task's environment is \emph{correct} when it runs the original game and implements the task's start, progress and success faithfully, and a task is \emph{solved} when its environment is correct and the final code policy completes it from two seeds.

\noindent\textbf{Results.}
\Cref{fig:loop} splits the 145 tasks of each setting into solved, correct but unsolved, and incorrect.
The generic agent leaves 120 environments incorrect and solves one task, so the same model without our skills, tools and players cannot build faithful environments.
With our Builder, player reports cut the incorrect environments from 51 to 40 and raise the solved tasks from 34 to 46.
They work in two ways: defect reports expose bugs that the Builder's own tests missed, and observation requests give the \cp state that it cannot read itself (\cref{tab:loop-repairs-main}).
In \emph{There Is No Game}, for example, the \cp reported that progress stopped at 0.9, and after the repair all nine tasks are correct and solved, while without reports all nine remain wrong.
With GPT-6.1 Sol as the Builder, the full loop and the no-reports setting both give 136 correct environments, and reports still raise the solved tasks from 62 to 82.

\noindent\textbf{Cost.}
Reports add 20\% to the model cost per game, from \$17.7 to \$21.3.
Recording gameplay then costs the two players differently in kind (\cref{fig:recording}).
A \vp pays for every step: GPT-6.1 Sol costs \$54.7 per hour of gameplay.
A code policy pays mostly once: writing it with the same model costs \$10.8, after which it plays without model calls at \$0.13 per hour of gameplay.
It therefore becomes cheaper than the \vp after 12 minutes of gameplay, and our training data, with 40 minutes of code policy gameplay per game on average, lies well beyond this point.
A code policy also records data much faster, because the game runs in real time for it but pauses while the \vp decides.
In an hour of wall-clock time, the code policy records 7,775 harness steps, whereas the \vp records only 354.

\subsection{Main results}
\label{sec:exp-policy}
\label{sec:exp-train}

\input{tables/tab_leaderboard}
\input{figures/float_students}
\begin{figure}[t]
\centering
\includegraphics[width=\linewidth]{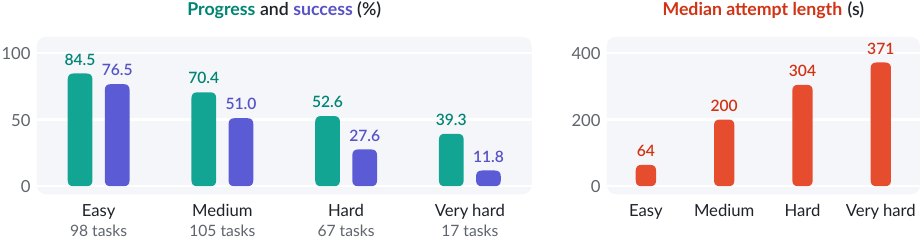}
\caption{\textbf{Human play by difficulty tier} on \bench. Left: progress and success in \%, averaged over the tasks of each tier, with the number of tasks under each tier. Right: median length of all attempts in seconds.}
\label{fig:human-tiers}
\end{figure}

\Cref{tab:players} compares our trained models with human players, closed- and open-source \vp{}s and game-specific models on both benchmarks.

\noindent\textbf{\bench is more challenging than \gw.}
Every model evaluated on both benchmarks except \mbox{Qwen3.8-Flash} scores lower on \bench, by 12.8 progress and 7.6 success points on average.
Even human players, 15 first-time participants who play in real time with the same instructions as the \vp{}s (\cref{supp:human}), reach only 69.4\% progress and 51.8\% success, against 82.6\% and 77.1\% on \gw.
A likely reason is that \gw uses games with available source code, whereas we admit any game that can be downloaded.

\noindent\textbf{Difficulty tiers match human play.}
The Builder labels every task as easy, medium, hard or very hard (\cref{supp:split}), and human play follows this ordering (\cref{fig:human-tiers}): success falls from 76.5\% on easy tasks to 11.8\% on very hard ones, and the median attempt grows from about one minute to six.

\noindent\textbf{Training roughly doubles \bench progress and transfers to \gw.}
With thinking on, \gnine reaches 31.2\% progress and 15.0\% success on \bench and 53.1\% and 29.4\% on \gw, the best among open-source models under 10B parameters and above the game-specific PLAYER-1~\cite{ouyang2026player1} and GameScaling~\cite{zhang2026gamescaling} models.
Compared with the Qwen3.5 base models, both trained models roughly double their \bench progress and gain more than 20 points on \gw, although \gw shares no games, tasks or success checks with our collection.
Across four evaluation seeds with thinking off, the standard deviation of progress stays within 2.5 points (\cref{tab:repeat}), far below these gains.
On \gw, \gnine even surpasses general-purpose models such as Qwen3.8-Flash, GLM-5.3-Flash, GPT-6~Luna and Claude Haiku~5.5, and comes close to \mbox{Claude Sonnet 5.5}.
On \bench, however, the strongest closed-source models reach about 60\% progress, leaving much room for open models to improve.

\noindent\textbf{Training makes thinking useful.}
\Cref{fig:students} evaluates each model with thinking on and off, using the same parameters.
Before training, thinking has small and mixed effects: it changes progress by $+3.2$ and $-1.9$ points on \bench for the 4B and 9B models, and by $-0.4$ and $+1.3$ on \gw.
After training, it adds 7.7 and 2.6 points on \bench and 14.6 and 19.2 on \gw.
Thinking thus consistently improves performance once the models are trained on our gameplay, most of all on the unseen games of \gw.

\subsection{Ablations and analysis}
\label{sec:exp-ablation}

\input{figures/float_scaling}

\input{tables/tab_ablation}

\Cref{tab:ablation,fig:scaling} vary the data source, the sampling weights, the amount of data and the number of training steps for both model sizes. We run all ablations with thinking off, which substantially reduces inference cost and allows a broader set of comparisons.

\noindent\textbf{Data source.}
Each kind of player improves the base models on its own, and the two help in different ways.
On \bench, \vp data is far more sample-efficient: with 0.19M examples it raises the 4B model from 12.7\% to 25.1\% progress, against 17.3\% for 1.37M code policy examples, since it matches the evaluation format, with paused steps from screenshots, and keeps the model's notes and any native thinking.
On \gw, code policy data alone gives the best model at both sizes, with 36.8\% and 38.0\% progress, likely because its precisely timed key presses and holds transfer to games outside our collection.
Combining both sources gives the best 9B model on \bench, with 28.6\% against 25.6\% and 20.1\% for single sources, while the 4B combination falls between them.

\noindent\textbf{Progress weighting.}
At an equal number of examples, progress-weighted SFT outperforms uniform sampling from the same recordings for both model sizes and on both benchmarks.
It raises \bench progress from 20.9\% to 23.4\% for the 4B model and from 23.9\% to 28.6\% for the 9B model, and \gw progress from 33.5\% to 34.9\% and from 31.1\% to 33.9\%.
Its credits, penalties and game balancing (\cref{sec:distill}) focus training on the few actions that move a task forward.

\noindent\textbf{Data volume and training steps.}
More training data keeps helping (\cref{fig:scaling}, left): even 100K examples improve on the base models, and growing to all 1.56M raises \bench progress from 21.2\% to 23.4\% for 4B and from 21.2\% to 28.6\% for 9B.
Over training steps, \bench progress keeps rising until the final checkpoint, while \gw progress plateaus after about 3K steps (\cref{fig:scaling}, right).

\subsection{Case studies on held-out tasks}
\label{sec:exp-case}

\input{figures/float_case}

An important behavior that training teaches is precise timing control.
In the held-out platformer \emph{Give Up} (\cref{fig:case}, top left), how long the direction keys are held largely decides how far the character runs and jumps.
Within 100 steps, Qwen3.5-4B reaches 12.3\% progress, while \gfour reaches 92.5\%.
\gfour sets an explicit hold duration on 47\% of its key presses, against 1\% for its base model.
With identical harness instructions for both models, the model learns timing control from our data, in which code policy recordings are cut into timed holds and \vp{}s use holds routinely.

The other three tasks in \cref{fig:case} come from different genres.
In \emph{Alfonzo's Bowling Challenge}, the task is to destroy forty bricks with a ball steered by the arrow keys.
Qwen3.5-4B sets a hold duration on 2 of its 98 key presses, alternates between left and right, and destroys two bricks.
\gfour sets a duration on 69 of its 71 key presses, waits on 29 steps to follow the ball and destroys 39 bricks within 100 steps.
In \emph{Kingdom Rush}, the task is to build four towers.
Qwen3.5-4B clicks and presses keys for 100 steps without building a tower, and 11 of its steps are invalid.
\gfour opens the build menu on grass tiles beside the road, waits for gold when it cannot pay for the next tower, and builds the fourth tower at step 56.
In the card game \emph{Forward}, the knight moves by being dragged onto an adjacent card.
Qwen3.5-4B mixes 34 clicks with 58 drags, and its progress stops at 20\% after step 23.
\gfour drags the knight from card to card and clears the area in 78 steps.
Before some of its moves it hovers over a card to read it.
\Cref{fig:qual} in \cref{supp:qual} adds the task instructions and the initial frames.

%% file: figures/float_loop.tex
\begin{figure}[t]
\centering
\begin{minipage}[t]{0.49\linewidth}\vspace{0pt}
\centering
\includegraphics[width=\linewidth]{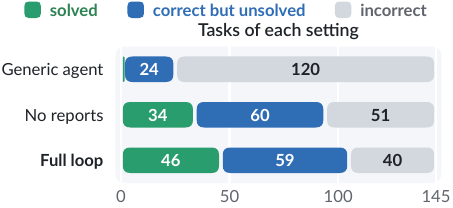}
\caption{\textbf{Task outcomes of co-development} on 20 games and 145 frozen tasks, with a GPT-6.1 Terra Builder and GPT-6.1 Sol players.}
\label{fig:loop}
\end{minipage}\hfill
\begin{minipage}[t]{0.49\linewidth}\vspace{0pt}
\centering
\includegraphics[width=\linewidth]{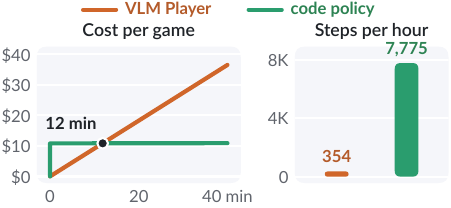}
\caption{\textbf{Recording gameplay with GPT-6.1 Sol.} The cost of recording gameplay for one game (\textit{left}) and the harness steps recorded per hour of wall-clock time (\textit{right}), by the \vp and by the code policy.}
\label{fig:recording}
\end{minipage}
\end{figure}

%% file: tables/tab_codev_repairs_main.tex
\begin{table}[t]
\centering\compacttable[7.2]
\begin{tabularx}{\linewidth}{@{}l>{\raggedright\arraybackslash}X>{\raggedright\arraybackslash}Xrr@{}}
\toprule
\rowcolor{tblheadbg}\tblhead{Game} & \tblhead{Report} & \tblhead{Change in the full loop} & \tblhead{Correct} & \tblhead{Solved} \\
\midrule
There Is No Game & \cp: progress stays at 0.9 after the letter O falls & Hooks each game variable when it appears, so later story events are recorded & 0/9 & 0/9 \\
Dungeons \& Degenerate Gamblers & \vp: the table stays empty after the Gambler is chosen; \cp: cards and buttons cannot be read & Cancels pending room animations before freeing the cards; adds cards, buttons and battle state & 6/7 & 0/0 \\
Dome Romantik & \vp: the upgrade list runs below the screen & Keeps five upgrades on screen and scrolls the list with the selection & 5/6 & 4/0 \\
Where the Water Flows & \cp: terrain and objects cannot be read & Adds terrain, platforms, mechanisms and movable objects & 7/7 & 0/4 \\
Strike Force Heroes & \cp: the player, soldiers and flags cannot be read & Adds the player, visible units, flags and pickups & 5/5 & 0/4 \\
\bottomrule
\end{tabularx}
\par
\caption{\textbf{Selected reports and repairs with the GPT-6.1 Terra Builder.} Correct and Solved give the number of tasks without reports / with the full loop. \Cref{tab:loop-repairs} lists all reports.}
\label{tab:loop-repairs-main}
\end{table}

%% file: tables/tab_leaderboard.tex
\begin{table}[t]
\centering\compacttable[7.4]
\setlength{\tabcolsep}{7pt}
\begin{tabular}{@{}l rr @{\hspace{18pt}} rr@{}}
\toprule
\rowcolor{tblheadbg} & \multicolumn{2}{c}{\tblhead{\bench}} & \multicolumn{2}{c}{\tblhead{\gw}} \\
\cmidrule(lr){2-3}\cmidrule(l){4-5}
\rowcolor{tblheadbg}\tblhead{Model} & \tblhead{Progress} & \tblhead{Success} & \tblhead{Progress} & \tblhead{Success} \\
\midrule
Human players & 69.4 & 51.8 & 82.6$^\dagger$ & 77.1$^\dagger$ \\
\addlinespace[2pt]
\tblsection{5}{Closed-source models}
GPT-6.1 Sol & \tblbold{62.7} & 39.0 & 64.3 & 44.1 \\
\hspace{1em}medium effort & 62.1 & 40.1 & -- & -- \\
GPT-6.1 Terra & 60.4 & 36.2 & \tblbold{69.2} & \tblbold{49.4} \\
GPT-6 Luna & 38.5 & 20.9 & 51.7 & 27.1 \\
Claude Opus 5.5 & 58.9 & 38.3 & 65.5 & 47.6 \\
\hspace{1em}medium effort & 62.0 & \tblbold{41.1} & 65.6 & 47.6 \\
Claude Sonnet 5.5 & 47.9 & 28.9 & 54.7 & 31.8 \\
Claude Haiku 5.5 & 40.9 & 23.7 & 47.0 & 26.5 \\
Gemini-3.8-Flash & 30.3 & 15.7 & 57.3 & 33.5 \\
\addlinespace[2pt]
\tblsection{5}{Open-source models}
DeepSeek-V4.1-Flash & 39.5 & 23.3 & \tblbold{57.2} & \tblbold{34.1} \\
GLM-5.3-Flash & 38.2 & 22.0 & 48.6 & 25.9 \\
Qwen3.8-Flash & \tblbold{45.1} & \tblbold{24.7} & 44.4 & 22.4 \\
Qwen3.8-27B & 34.6 & 18.5 & 47.7 & 25.3 \\
Qwen3.5-35B-A3B & 17.9 & 7.3 & 34.3 & 15.3 \\
Qwen3.5-4B & 15.9 & 5.6 & 26.9 & 8.2 \\
Qwen3.5-9B & 14.3 & 5.2 & 30.0 & 13.5 \\
\addlinespace[2pt]
\tblsection{5}{Game-specific models}
PLAYER-1-9B~\cite{ouyang2026player1} & 19.0 & 7.0 & 34.5$^\S$ & 15.9$^\S$ \\
PLAYER-1-35B-A3B~\cite{ouyang2026player1} & \tblbold{19.2} & \tblbold{9.1} & \tblbold{38.8$^\S$} & \tblbold{17.3$^\S$} \\
GameScaling-9B-S3~\cite{zhang2026gamescaling} & 9.6 & 3.1 & 23.6 & 7.1 \\
\addlinespace[2pt]
\tblsection{5}{Published \gw results}
GPT-5.6 Sol (CU)$^\ddagger$ & -- & -- & \tblbold{62.9} & \tblbold{40.0} \\
Gemini-3-Flash-Preview$^\ddagger$ & -- & -- & 41.9 & 21.2 \\
GPT-5.2$^\ddagger$ & -- & -- & 40.6 & 20.6 \\
Seed-1.8 (CU)$^\ddagger$ & -- & -- & 39.8 & 20.0 \\
Claude-Sonnet-4.6$^\ddagger$ & -- & -- & 39.3 & 20.6 \\
Kimi-K2.5$^\ddagger$ & -- & -- & 37.4 & 18.8 \\
\addlinespace[2pt]
\tblsection{5}{Ours}
\rowcolor{tbloursbg}\gfour & 31.1 & 14.6 & 49.5 & 27.6 \\
\rowcolor{tbloursbg}\gnine & \tblbold{31.2} & \tblbold{15.0} & \tblbold{53.1} & \tblbold{29.4} \\
\bottomrule
\end{tabular}
\caption{\textbf{Benchmark results on \bench and \gw.} Progress and success in \%. Bold marks the best of each group, and our models are shaded. All models play with thinking on, and API models at low reasoning effort unless marked as medium; \cref{fig:students} compares thinking on and off. Averages in the text use the low-effort rows of the models evaluated on both benchmarks. Unless marked, entries use our harness with 100 steps and one run per task, and game-specific models their own harnesses adapted to our games. $^\dagger$Expert player reported by \gw~\cite{arxiv260407429}. $^\ddagger$Published results with the \gw harness~\cite{arxiv260407429,han2026vista}; CU denotes its computer-use mode, the others its generalist mode. $^\S$Published results under the PLAYER-1 protocol with 400 steps and three trials per task~\cite{ouyang2026player1}.}
\label{tab:players}
\label{tab:leaderboard}
\end{table}

%% file: figures/float_students.tex
\begin{figure}[t]
\centering
\includegraphics[width=\linewidth]{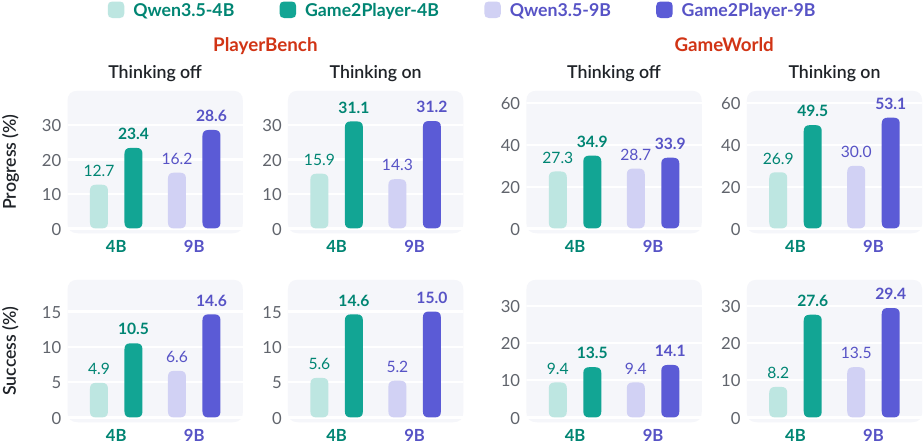}
\caption{\textbf{Base and \sys models with thinking on and off.} Progress (\textit{top}) and success (\textit{bottom}) in \% of each Qwen3.5 base model (light) and its \sys counterpart (dark) on both benchmarks.}
\label{fig:students}
\end{figure}

%% file: figures/float_scaling.tex
\begin{figure}[t]
\centering
\includegraphics[width=\linewidth]{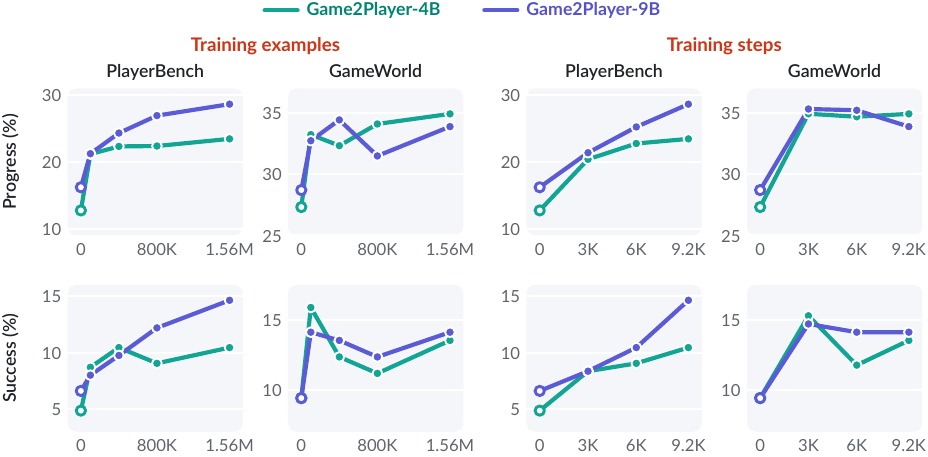}
\caption{\textbf{Training data and training steps.} Progress and success on \bench and \gw with 0, 100K, 400K, 800K and all 1.56M training examples (\textit{left}) and after 0, 3K, 6K and all 9.2K training steps (\textit{right}). Hollow points at zero denote the untrained Qwen3.5 base models.}
\label{fig:scaling}
\end{figure}

%% file: tables/tab_ablation.tex
\begin{table}[t]
\centering\compacttable[7.8]
\begin{tabularx}{\linewidth}{@{}ll*{4}{>{\raggedleft\arraybackslash}X}@{}}
\toprule
& & \multicolumn{2}{c}{\tblhead{\bench}} & \multicolumn{2}{c}{\tblhead{\gw}} \\
\cmidrule(lr){3-4}\cmidrule(l){5-6}
\rowcolor{tblheadbg} & \tblhead{Setting} & \tblhead{Progress} & \tblhead{Success} & \tblhead{Progress} & \tblhead{Success} \\
\midrule
\tblsection{6}{Data source}
\multirow{4}{*}{4B} & Base & 12.7 & 4.9 & 27.3 & 9.4 \\
& \cp only & 17.3 & 6.3 & \tblbold{36.8} & \tblbold{14.7} \\
& \vp only & \tblbold{25.1} & \tblbold{11.5} & 35.5 & 13.5 \\
& Both & 23.4 & 10.5 & 34.9 & 13.5 \\
\cmidrule(l){2-6}
\multirow{4}{*}{9B} & Base & 16.2 & 6.6 & 28.7 & 9.4 \\
& \cp only & 20.1 & 8.4 & \tblbold{38.0} & \tblbold{14.1} \\
& \vp only & 25.6 & 11.5 & 35.0 & 12.9 \\
& Both & \tblbold{28.6} & \tblbold{14.6} & 33.9 & \tblbold{14.1} \\
\midrule
\tblsection{6}{Sampling}
\multirow{2}{*}{4B} & Uniform & 20.9 & 8.4 & 33.5 & 11.8 \\
& Progress-weighted & \tblbold{23.4} & \tblbold{10.5} & \tblbold{34.9} & \tblbold{13.5} \\
\cmidrule(l){2-6}
\multirow{2}{*}{9B} & Uniform & 23.9 & 10.1 & 31.1 & 10.6 \\
& Progress-weighted & \tblbold{28.6} & \tblbold{14.6} & \tblbold{33.9} & \tblbold{14.1} \\
\bottomrule
\end{tabularx}
\par
\caption{\textbf{Data-source and sampling ablations.} Progress and success in \%. Uniform sampling draws as many examples without weighting or balancing.}
\label{tab:ablation}
\end{table}

%% file: figures/float_case.tex
\begin{figure}[t]
\centering
\includegraphics[width=\linewidth]{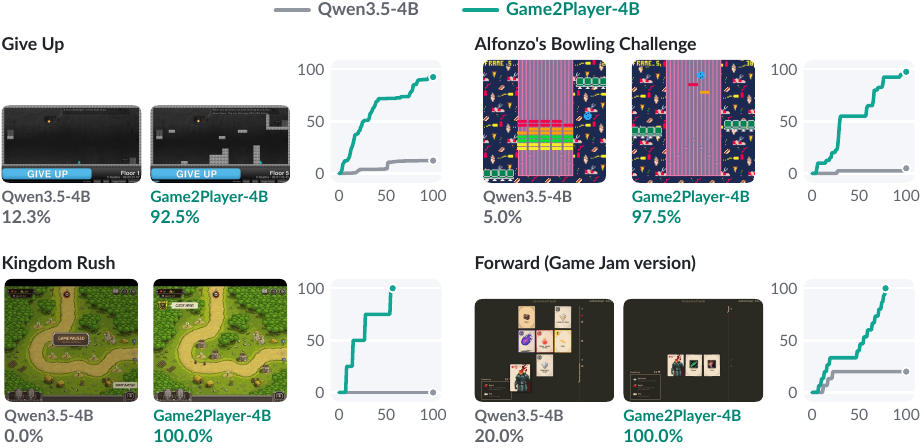}
\caption{\textbf{Qwen3.5-4B and \gfour on four held-out \bench tasks}, with thinking on and at most 100 steps. Each panel shows the final frame of each model with its final progress, and the best progress reached after each step.}
\label{fig:case}
\end{figure}

%% file: sections/5_conclusion.tex
\section{Conclusion}
\label{sec:conclusion}

\sys is a scalable pipeline that co-develops verifiable game environments with the players that use them, whose reports catch defects the Builder's own tests miss.
The same loop scales up gameplay experience through cheap code policy reruns and \vp episodes that keep notes and thinking.
Distilled with progress-weighted SFT, this experience roughly doubles the \bench progress of Qwen3.5 models and transfers to \gw.
\paragraph{Limitations.}
Accepted environments are not guaranteed to be correct.
An environment is accepted once it passes the Builder's evaluation, the scripted gate and the players' reports, without a full human audit.
The audit of the co-development study finds that 72\% of the tasks built by the GPT-6.1 Terra Builder and 94\% of those built by the GPT-6.1 Sol Builder are implemented correctly, and human review of all 287 \bench tasks found no environment or scoring errors (\cref{supp:human}).
The pipeline covers single-player web games that run offline in a browser, and online, multiplayer and large commercial games add licensing and runtime constraints that we do not address.
A \vp plays with the game paused at every decision, so our scores do not measure play under real-time latency.
The Builder, the players and many baselines are proprietary models whose versions change over time, so rebuilding the collection with other models may give different environments.
We discuss the broader impact in \cref{supp:impact}.

%% file: sections/X_suppl.tex
\clearpage
\beginappendix
\renewcommand{\theHsection}{supp.\Alph{section}}
\crefname{appendix}{Appendix}{Appendices}\Crefname{appendix}{Appendix}{Appendices}
\crefalias{section}{appendix}\crefalias{subsection}{appendix}

\section{Environment construction}
\label{supp:construction}

\paragraph{Scout.}
The Scout searches game portals such as \textit{itch.io} and other public source repositories.
For each candidate it checks whether the game can be played alone and offline, whether its goals split into short tasks, whether it can be restarted reliably, and how its state could be read, for example from named variables in the source.
It records evidence of play and the game's license, and recommends to build, reserve or skip the game.

\paragraph{Builder phases.}
The Builder runs four phases.
\phase{Discover} inspects feasibility and proposes tasks grounded in the source or in play.
\phase{Implement} packages the offline game and implements the \gapi, a JavaScript module loaded with the game, the control guide, and the start routine, progress and success of every task, editing only a fixed set of paths.
\phase{Evaluate} plays every task start from the player's view, then tests real controls, state changes and task definitions without editing files.
It labels each task's start, progress and success as verified, pending or unsupported.
A Builder pass requires that every task still matches the frozen plan, every start is verified, no progress or success definition is unsupported, and every verified progress definition cites a recorded change of its state variable.
It also requires at least one state variable read from the game rather than estimated from the screen, and a check from the player's view that the screen shows no defect and that every control does what the control guide says.
\phase{Revise} diagnoses the recorded evidence of the last evaluation before applying a bounded fix.
The evaluation instructions state that the presence of an interface or a zero exit code is not evidence and that tasks should not be solved merely to pass evaluation.

\paragraph{Retired tasks and unsuitable games.}
A frozen task that the game itself cannot complete, or whose start, progress or success still cannot be measured faithfully after a bounded attempt, is retired with its reason and a record that shows it.
A task that is long or hard stays in the plan.
At most a third of the plan may be retired.
A game that would need more retired tasks, or whose progress could only be estimated from the screen, is unsuitable.

\paragraph{Engine skills.}
The pipeline uses 13 skills, namely the Builder skill, eight engine skills for HTML5 capture, Unity WebGL, Flash, PICO-8, GameMaker, Godot, Construct and Defold, and four policy skills for general policies, canvas user interfaces, PICO-8 and Unity.
The Builder selects an engine skill from the detected engine, and the \cp selects the relevant policy guidance.
Each skill is a short instruction file with references to engine tools, and \cref{fig:skill} shows the skill for Godot.

\begin{figure}[t]
\centering
\begin{promptbox}{rolebuilder}{Engine skill for Godot}
\promptfile{figures/prompt/skill_godot.txt}
\end{promptbox}
\caption{\textbf{An engine skill.} The Builder receives this skill once it has identified a Godot game. The linked workflow and tools are part of the skill.}
\label{fig:skill}
\end{figure}

\paragraph{State contract and scripted gate.}
The state returned by the \gapi contains a schema version, the game identifier, wall-clock and game time, a status, whether the game is actionable, a terminal outcome, a game-specific state block and numeric metrics.
The scripted gate checks these types, an actionable start, numeric score paths for every task, the terminal outcome domain, task definitions against the frozen plan, and finite targets above their start values.
For each task start it loads the game with the external network blocked and a fixed seed, and then tests two resets.

\paragraph{\cp procedure and acceptance.}
The \cp runs \phase{Discover}, \phase{Implement}, \phase{Evaluate}, \phase{Diagnose} and \phase{Revise}.
It inspects the visible information and the player observation, writes one code policy per game, records bounded headless episodes, separates policy failures from environment defects and retests after each revision.
The \cp collects the player observation with a probe that it writes for each game.
Every field of the probe comes from a named source, such as a game object or a page element, and the probe may read anything a person could see on the screen but no hidden information.
When a visible quantity cannot be read without changing the game, the \cp files an observation request, and the Builder adds the quantity to the player observation.
Every task needs at least two recorded attempts, and every task reported as completed needs at least two successful episodes of the current code policy.
For a game with $n>1$ tasks, a code policy is accepted if it completes at least $\max(2,\lfloor n/4\rfloor+1)$ tasks, and a single-task game requires that task.

\paragraph{Reports and repair.}
\Cref{tab:report-types} lists what each player may report.
The reports of a game start a new Builder run, which goes through all four phases with the frozen task plan.
In \phase{Evaluate}, the Builder gives each report one of three verdicts.
A report is fixed when the Builder found the defect, repaired it and checked the affected behavior through player inputs.
It is disproved when a check from the player's side explains or contradicts the reported failure.
It is unresolved when neither holds, and the \cp then examines it while writing its policy.
An environment is accepted only when every report is fixed or disproved.
Until then it stays pending, and no gameplay is recorded in it.
After an observation request, the \cp checks each added field against the screen before its policy relies on it.
Failures that the harness detects by itself, such as a game that does not start or reset, return to the Builder in the same way, and the affected tasks are played again after the repair.

\paragraph{Agent prompts.}
Each agent works from a prompt that states its scope and the rules its results must satisfy.
\Cref{fig:agent-prompts} quotes rules from the prompts of the Scout, the Builder and the \cp.

\begin{figure*}[tp]
\centering
\begin{promptbox}{rolescout}{Scout}
\promptfile{figures/prompt/scout.txt}
\end{promptbox}
\begin{promptbox}{rolebuilder}{Builder}
\promptfile{figures/prompt/builder.txt}
\end{promptbox}
\begin{promptbox}{rolecoder}{Coding Player}
\promptfile{figures/prompt/coder.txt}
\end{promptbox}
\caption{\textbf{Rules quoted from the agent prompts.} The Scout selects games, the Builder freezes the tasks, measures progress toward the objective, plays from the player's view and checks player reports against recordings, and the \cp reads only its probe and asks only for observable state. Grey brackets mark omitted text.}
\label{fig:agent-prompts}
\end{figure*}

\input{tables/supp_report_types}

\section{Coding Player evaluation}
\label{supp:policy}

\paragraph{Protocol.}
For each \bench game, the pipeline built a code policy as in \cref{sec:codev}.
The games were built at different times with different coding agents as the \cp, listed in \cref{tab:policy-authors}.

\input{tables/supp_policy_authors}
Each task is played once in real time from the same seed, with up to 600 seconds of gameplay.
Policies exist for 245 of the 287 tasks across 43 games, and the remaining 42 tasks count as failures with zero progress.

\paragraph{Per-game and per-genre results.}
\Cref{tab:policy-per-game} gives the progress of the \cp and of the best of the four compared \vp{}s on each game, and \cref{fig:policy} plots the two against each other. The largest mean progress gaps occur in racing, with 100.0 against 18.2, and rhythm games, with 80.0 against 33.1.

\input{figures/float_policy}

\input{tables/supp_policy_per_game}

\paragraph{Trajectory diversity.}
We compare two runs $a$ and $b$ of the same task by their progress and by the keys they press.
Let $p_a(t)$ be the best progress that run $a$ has reached by time $t$ and $T$ the duration of the longest run of the task.
The progress distance is
\begin{equation}
D_{\mathrm{prog}}(a,b)=\frac{1}{T}\int_0^T\bigl|p_a(t)-p_b(t)\bigr|\,dt.
\end{equation}
A progress curve rises from 0 to at most 1, so if the progress that a run never reaches is placed at time $T$, the curve is a cumulative distribution of when progress arrives, and $D_{\mathrm{prog}}$ is the earth mover's distance~\cite{rubner2000earth} between these distributions divided by $T$.
Let $k_a$ be the sequence of keys that run $a$ presses, in order.
The key distance is
\begin{equation}
D_{\mathrm{key}}(a,b)=\frac{\mathrm{Lev}(k_a,k_b)}{\max(|k_a|,|k_b|)},
\end{equation}
where $\mathrm{Lev}$ is the Levenshtein distance~\cite{levenshtein1966binary}, the smallest number of insertions, deletions and substitutions of keys that turn one sequence into the other.
It is 0 for identical sequences and 1 for sequences without any common key.
Both distances are averaged over all pairs of runs of a task, then over the tasks of a game and then over games.
$D_{\mathrm{prog}}$ covers all 1,348 tasks in 330 games with repeated code policy runs, and $D_{\mathrm{key}}$ the 962 tasks in 248 games whose runs press keys, since clicks alone carry no key sequence.
Two runs of the same code policy differ by $D_{\mathrm{prog}}=7.3\%$ and $D_{\mathrm{key}}=0.28$.

\begin{figure*}[t]
\centering
\includegraphics[width=\linewidth]{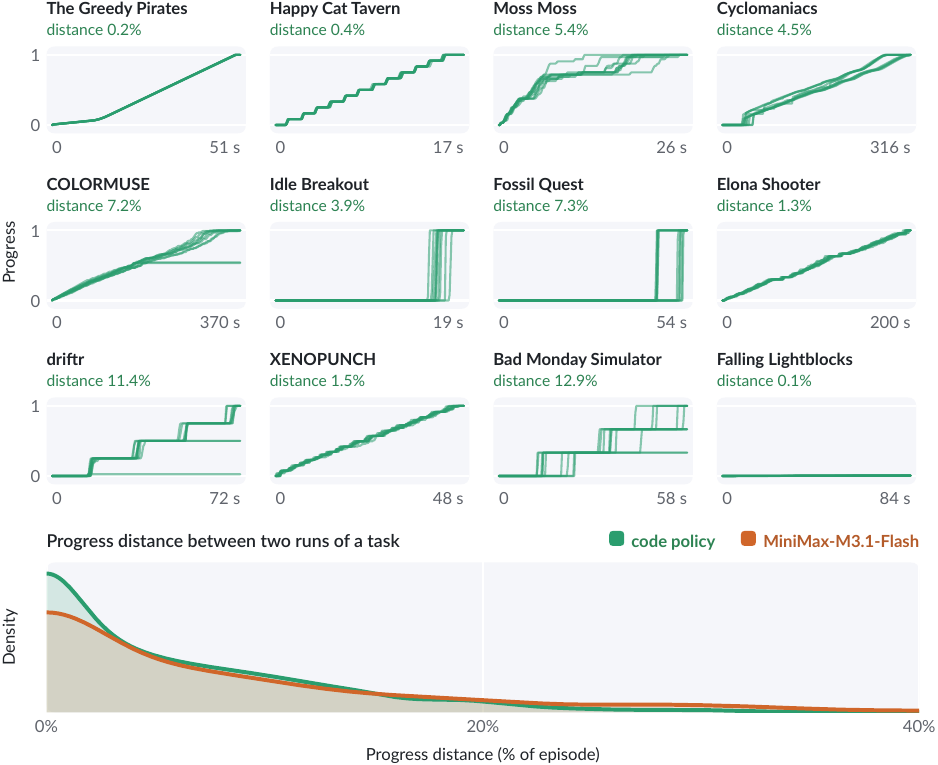}
\caption{\textbf{Progress of repeated code policy runs.} Top, progress of the ten runs of the same code policy on each of twelve tasks drawn at random from the tasks with ten runs, with the progress distance $D_{\mathrm{prog}}$ of each task. Bottom, the density of $D_{\mathrm{prog}}$ over tasks for the code policies and for repeated runs of MiniMax-M3.1-Flash.}
\label{fig:rerun-progress}
\end{figure*}

\begin{figure*}[t]
\centering
\includegraphics[width=\linewidth]{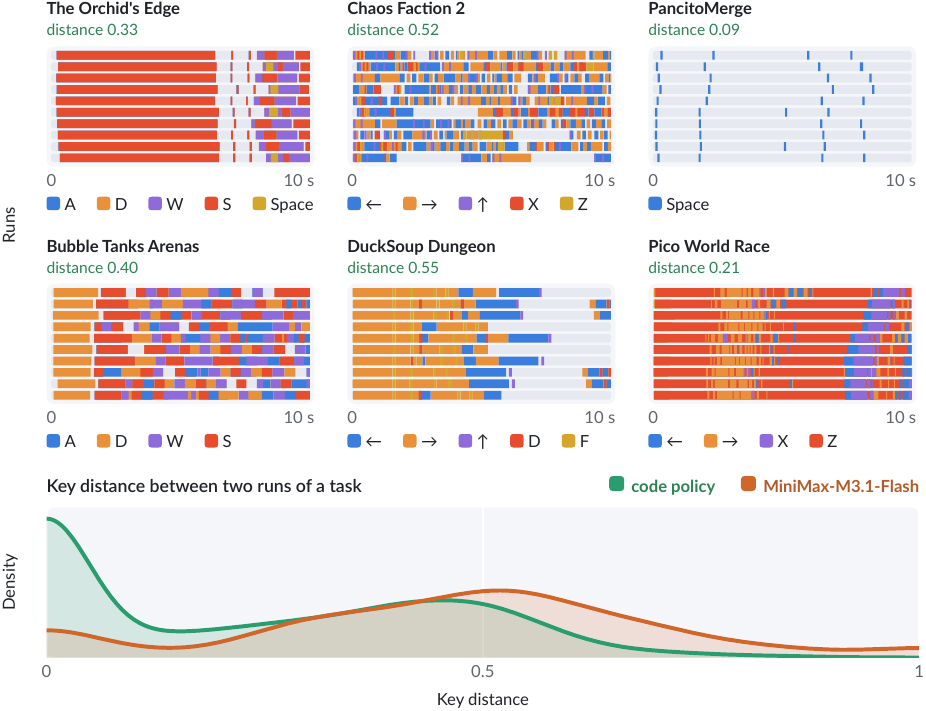}
\caption{\textbf{Key presses of repeated code policy runs.} Top, the key presses in the first ten seconds of the ten runs of the same code policy on each of six tasks drawn at random from the tasks whose runs press different keys, one lane per run and one colour per key, with the key distance $D_{\mathrm{key}}$ of each task. Bottom, the density of $D_{\mathrm{key}}$ over tasks for the code policies and for repeated runs of MiniMax-M3.1-Flash.}
\label{fig:rerun-keys}
\end{figure*}

As a baseline we use the repeated runs of MiniMax-M3.1-Flash, a \vp with low thinking and temperature 1, that were recorded with six runs per task over three seeds.
We compare its runs with different seeds, as every code policy rerun uses a new seed.
Over 346 tasks in 73 games it gives $D_{\mathrm{prog}}=10.2\%$, and over the 263 of these tasks that use keys it gives $D_{\mathrm{key}}=0.48$.
Reruns of a code policy thus differ in progress and in keys on a scale comparable to a \vp that samples every action.
\Cref{fig:rerun-progress,fig:rerun-keys} show progress curves and key presses of repeated code policy runs on randomly chosen tasks, together with the density of both distances over tasks.

\section{Training data and recipe}
\label{supp:training}

\paragraph{Data sources.}
\Cref{tab:data-sources} breaks down the training data by player and model.
Code policies provide 1,372.1K action-only examples, while the \vp{}s provide 191.5K examples with recorded notes or thinking.
Among the latter, 169.1K contain a note without thinking and 22.5K contain thinking.
The collection grew while we recorded gameplay, so the recordings cover 396 of the 518 training games, with 2,251 tasks.

\input{tables/supp_data_sources}

\begin{figure*}[t]
\centering
\includegraphics[width=\linewidth]{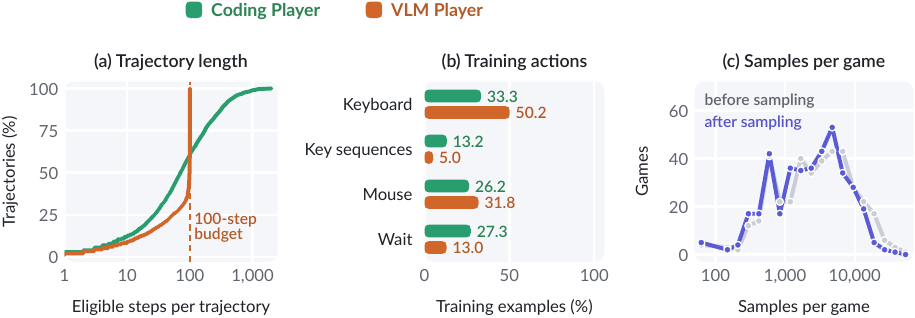}
\caption{\textbf{Characteristics of the training data.} (a) Cumulative fraction of trajectories by the number of eligible steps after segmentation and filtering, before resampling. The VLM Player has a 100-step budget. (b) Action composition of the final training examples. Individual keyboard actions, key sequences, mouse actions and waits are counted separately. (c) Distribution of samples per game before and after resampling. Each point counts the games in a sample-count interval, with neighboring interval counts connected by lines. Intervals use a logarithmic scale.}
\label{fig:data-stats}
\end{figure*}

\paragraph{Trajectory characteristics.}
\Cref{fig:data-stats} shows the distribution of trajectory lengths, the actions used as training targets, and how sampling changes the contribution of each game.
Lengths count eligible steps before resampling, including explicit waits.
The median trajectory has 71 steps for code policies and 99 for \vp{}s.
Waits make up 27.3\% and 13.0\% of their respective training examples.
The largest source game contributes 3.02\% of eligible steps and 2.39\% of the sampled training examples.
\Cref{tab:training-outcomes} reports the outcomes of the retained recordings from which these examples are drawn.

\input{tables/supp_training_outcomes}

\paragraph{Recording.}
Both players run in the same browser runtime and are recorded by the same recorder.
The recorder logs each input when it is sent, so input that a game ignores or generates by itself is neither lost nor invented.
It samples the state probe and task progress on the same clock as the video, and for \vp{}s it also stores every request, response and reasoning trace.

\paragraph{Segmentation and filtering.}
Held inputs are snapped to the nearest of 26 durations between 0.05 and 7.8 seconds.
Presses shorter than 0.075 seconds become taps.
Up to five consecutive key-state segments, each at most 0.25 seconds long, separated by at most 0.2 seconds and spanning at most 0.8 seconds, become one chained action, and longer holds are split at 7.8 seconds.
An idle gap between actions becomes a run of waits of at most 1 second each, and a remainder of 0.2 seconds or less is dropped.
Each action is paired with the frame at its start, and the preceding steps form its history as in the harness.
Episodes are kept when they have a valid outcome, a probed video and at least 90\% state coverage, and duplicates of the same source, game, task and seed are removed.
All \cp episodes of a task are excluded if no policy ever completed that task.

\paragraph{Effect of preprocessing.}
The recordings produce 2,088.5K candidate steps after action segmentation and wait insertion.
Basic validity filtering removes 13.6K invalid action targets.
Weighting assigns zero weight to 1.2K actions at failed endings and 19.6K waits.
Of the remaining positive-weight steps, weighted sampling leaves out 379.1K actions and 243.5K waits, then adds 132.2K repeated uses of actions.
This yields 1,563.6K training examples, comprising 1,164.9K actions and 398.8K waits (\cref{tab:preprocessing-counts}).

\input{tables/supp_preprocessing}

\paragraph{Sampling weights.}
A material progress event is a positive increment at least as large as both 0.02 and the 75th percentile of positive increments, or a crossing into a new progress decile, with events at least 0.35 seconds apart.
\Cref{tab:weighting} lists the weighting stages behind \cref{eq:weight}, and \cref{fig:weights} illustrates them.
Repeated-action penalties affect 80.9K action targets, and timeout penalties affect 7.5K.
Game balancing assigns an aggregate action budget of 1,165.0K uses to 1,413.0K valid action targets.
These weights determine the subsequent resampling rather than removing one step for every penalty.

Each weight $w_t$ becomes an integer multiplicity
\begin{equation}
m_t=\lfloor w_t\rfloor+\mathbf{1}\bigl[u_t<w_t-\lfloor w_t\rfloor\bigr],
\label{eq:pwsft}
\end{equation}
where $u_t\in[0,1)$ is a reproducible pseudo-random number, so that an example appears $w_t$ times in expectation.
Let $n_j$ be the number of eligible action targets of game $j$, excluding waits, and $M$ the median of these counts across games.
The budget of game $j$ is
\begin{equation}
B_j=\begin{cases}
n_j, & n_j\leq M,\\
\min\!\bigl(n_j,\operatorname{round}(M^{1-\alpha}n_j^{\alpha})\bigr), & n_j>M,
\end{cases}
\end{equation}
with $\alpha=0.8445$, which reduces the contribution of large games while keeping the full budget of smaller ones.
We set $\alpha$ by hand for this balance and did not tune it on evaluation results.

\input{tables/supp_weighting}

\paragraph{Notes, thinking and supervision.}
A short note explains the chosen action in the \texttt{note} field of the action response, while native thinking is the longer text between \texttt{<think>} and \texttt{</think>} before that response.
These are the two components $n_t$ and $r_t$ of $z_t$ in \cref{eq:traj}.
Only notes accompany earlier actions in the input history, and each target contains the thinking generated for its own step, when available.

Code policy recordings contain neither kind of text.
Their targets have an empty thinking prefix with zero loss weight, followed by the action response with normal loss weight.
For \vp recordings, the action response and the recorded note receive supervision, together with the native thinking when it was recorded.
Missing notes are left empty rather than generated after the episode.

Both inference modes use the same trained parameters.
With thinking on, the chat template opens the thinking span and the model generates it before the action response.
With thinking off, the template supplies an empty thinking span and generation begins with the action response.
The difference is an inference setting rather than a separate fine-tuning stage.

\paragraph{Training recipe.}
\gfour and \gnine use the same recipe and differ only in the base model.
Full fine-tuning keeps the vision encoder and aligner frozen and uses BF16, learning rate $10^{-5}$ with cosine decay and 3\% warmup, one epoch, and packed sequences of at most 25,000 tokens.
Training runs on eight 80\,GB GPUs with two packed sequences per device and three gradient-accumulation steps, for 9,237 steps.
The bottom row of \cref{fig:students} compares the success rates of the base and trained models with thinking on and off.

\section{Evaluation}
\label{supp:eval}

\paragraph{Prompt.}
The system prompt consists of a fixed opening, the game rules and the role and controls written by the Builder, the task instruction and the output format, which restricts actions to the game's allowed controls.
\Cref{fig:prompt} shows the system prompt and the user message of one step, with the parts written for each game and task replaced by placeholders.

\begin{figure*}[p]
\centering
\begin{minipage}[t]{0.495\linewidth}\vspace{0pt}
\begin{promptbox}{rolevlm}{System prompt}
\promptfile{figures/prompt/vlm_system.txt}
\end{promptbox}
\end{minipage}\hfill
\begin{minipage}[t]{0.495\linewidth}\vspace{0pt}
\begin{promptbox}{rolevlm}{User message}
\promptfile{figures/prompt/vlm_user.txt}
\end{promptbox}
\begin{promptbox}{rolevlm}{Answer}
\promptfile{figures/prompt/vlm_answer.txt}
\end{promptbox}
\end{minipage}
\caption{\textbf{Prompt of a \vp at one step,} with thinking on. Grey text in angle brackets stands for the game rules, controls and task instruction written by the Builder and for the content of earlier steps. The user message holds up to 50 previous steps for open-weight models and all previous steps for API models, the last six with their screenshots, and the current screenshot.}
\label{fig:prompt}
\end{figure*}

\begin{figure*}[t]
\centering
\includegraphics[width=\linewidth]{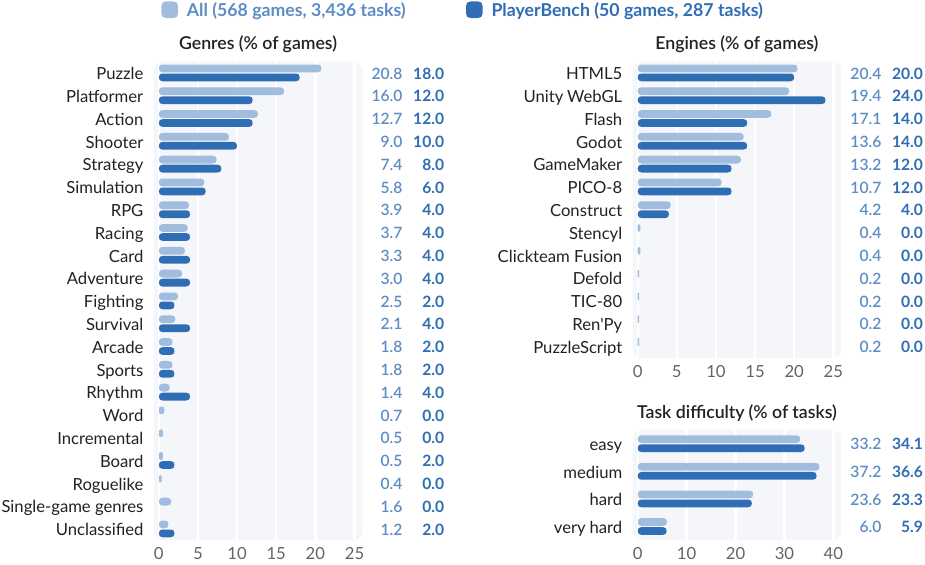}
\caption{\textbf{The collection and the held-out split in full.} Every genre and engine of the 568 games and the difficulty tiers of their 3,436 tasks, as shares of the whole collection (light) and of \bench (dark). The single-game genres are dress-up, interactive fiction, narrative, party, puzzle horror, runner, tower defense, typing and visual novel, and unclassified games fit none of the listed genres.}
\label{fig:collection-full}
\end{figure*}

\paragraph{History.}
For the models we train and the open-weight baselines, the context contains the current screenshot and up to 50 previous steps, each with its step number, action and note.
API models receive all previous steps.
Only the last six historical steps carry screenshots, so a request contains at most seven images.
Native thinking is not inserted into the history.

\paragraph{Action space.}
Each step returns one action, which may contain a timed key sequence.
Keyboard actions press one key, press several keys together, set held keys or play an ordered key sequence.
Pointer actions click, hold, move, drag and scroll, and the vocabulary also includes typing and waiting.
A key press or click without a duration is a 0.05-second tap, and durations are capped at 10 seconds.
Qwen models emit a JSON tool call with pointer coordinates normalized to 0--1000.

\paragraph{Budgets and failures.}
The step budget counts every model call, including calls without a valid action.
Reaching the budget without completing the task is a timeout that keeps the best progress reached.
Depending on the task, a terminal failure ends the episode or triggers a reset within the same budget.
A case counts as evaluated only with a valid recording, a usable outcome and at least 90\% state coverage, and infrastructure failures are rescheduled.

\paragraph{\bench composition.}
\label{supp:split}
\bench contains 50 games and 287 tasks, with 2 to 10 tasks per game.
The Builder assigns every task one of four difficulty tiers, and \bench has 98 easy, 105 medium, 67 hard and 17 very hard tasks.
By engine there are 12 Unity WebGL, 10 HTML5, 7 Flash, 7 Godot, 6 GameMaker, 6 PICO-8 and 2 Construct games.
By genre there are 9 puzzle, 6 action, 6 platformer, 5 shooter, 4 strategy and 3 simulation games, 2 games each of adventure, card, RPG, racing, rhythm and survival, and one game each of arcade, board, fighting, sports and other genres.
\Cref{fig:collection-full} compares the collection with the held-out split.

\paragraph{Game-specific models.}
\label{supp:leaderboard}
We evaluate PLAYER-1 on \bench and GameScaling-9B-S3~\cite{zhang2026gamescaling} on both benchmarks through their own prompt templates and action executors, with small adaptations to our games: the game rules, controls, allowed keys and task instructions come from our catalog, and their actions run in the same browser runtime.

\begin{table}[t]
\centering
\input{tables/supp_runs}
\end{table}

\paragraph{Variation across runs.}
We evaluate \gfour and \gnine with thinking off three more times, from seeds 42, 43 and 44, and \cref{tab:repeat} reports the mean and standard deviation over these runs and the run from seed 67.
The standard deviation stays within 2.5 points of progress on both benchmarks, well below the gains of 10.7 and 12.4 points over the base models on \bench with thinking off.
Differences of a few points between the settings of \cref{tab:ablation} are comparable to this variation, so we read them as trends.

\paragraph{Harness sensitivity.}
We evaluate six models on all 170 \gw tasks with our harness and with the \gw generalist harness, each with 100 steps (\cref{tab:harness-comparison}).
Our harness exposes keyboard and mouse commands with optional durations and timed key sequences, while the generalist harness maps game-specific semantic tools to predefined inputs.
Neither harness is consistently better.
Ours yields higher scores for Claude Opus 5.5 and GLM-5.3-Flash, and the generalist harness for GPT-6.1 Sol and Qwen3.8-Flash.
We therefore evaluate \gw with the same harness and settings as \bench, keeping its original tasks and evaluator.

\paragraph{History length.}
\Cref{tab:history} varies the text history and the number of historical screenshots for DeepSeek-V4.1-Flash and Qwen3.8-Flash on \bench.
The effect is mixed: a shorter history helps one model and hurts the other, so neither setting is consistently better.

\section{Co-development study}
\label{supp:loop}

\begin{table*}[tp]
\centering
\begin{minipage}{\linewidth}
\centering
\input{tables/tab_codev_games}
\end{minipage}

\vspace{14pt}
\begin{minipage}{\linewidth}
\centering
\input{tables/tab_codev_repairs}
\end{minipage}
\end{table*}

\begin{table*}[tp]
\centering
\begin{minipage}{0.6\linewidth}
\centering
\input{tables/tab_codev_cost}
\end{minipage}

\vspace{14pt}
\begin{minipage}{\linewidth}
\centering
\input{tables/tab_recording_rates}
\end{minipage}

\vspace{14pt}
\includegraphics[width=\linewidth]{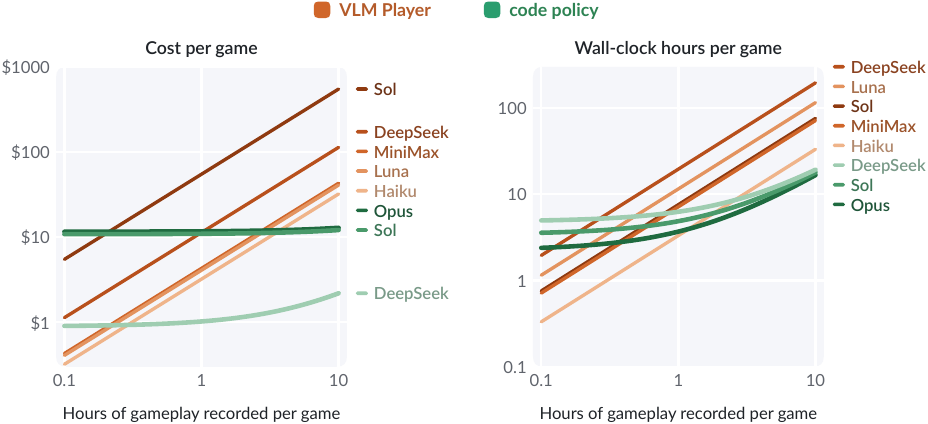}
\captionof{figure}{\textbf{Cost and wall-clock time of recording gameplay for one game, by model,} at the rates of \cref{tab:recording-rates}. The horizontal axis counts the time that passes in the game, and wall-clock time is the real time that recording takes, including the pauses while a \vp decides. Both axes are logarithmic.}
\label{fig:cost-models}
\end{table*}

\paragraph{Setup.}
The study uses 20 games from the collection, spread over seven engines, with their 145 task definitions frozen, so all settings implement the same goals.
In the full loop and the no-reports setting, the Builder builds each game once with its engine skills and tools.
The \vp then plays every task and the \cp writes a code policy until the first player report, and the run forks at that report.
In the full loop, the report returns to the Builder, which checks it against the recording and repairs the environment, and the players continue on the repaired environment, for up to two rounds of repair.
Each Builder run may revise its environment up to three times, the \cp may revise its policy up to 50 times, and the \vp plays each task for up to 100 steps from seed 67.
In the no-reports setting, no report reaches the Builder, and the \cp continues on the delivered environment with the same budget.
The two settings thus start from the same environment and differ only in the reports, and a game without any report ends identical in both.
In the generic-agent setting, a generic coding agent receives the original game, the frozen task requests and the basic interface to run the game, without project skills or engine-analysis tools.
A second generic agent then writes a code policy for the delivered environment, which it may not edit, and both agents use the Builder's model.
Each generic agent works in a single session of up to 90 minutes and may test its own work.
The final code policies of the full loop and the no-reports setting play every task from seeds 67 and 68 for up to 600 seconds, and the policy writer of the generic-agent setting records its own episodes.
A task is solved when its environment is correct and two of these episodes complete it.
We run the study with a GPT-6.1 Terra Builder and GPT-6.1 Sol players, as in \cref{sec:exp-env}, and repeat it with GPT-6.1 Sol for every agent, all at medium reasoning effort.

\paragraph{Judging environments.}
One auditor session per game, GPT-6.1 Sol at high reasoning effort, compares the environments of all settings with the original game files and the frozen task requests.
It traces each metric to native game state, inspects the recorded episodes and runs its own checks, and for each task it decides whether the environment runs the original game and whether its start, progress and success match the task.
Humans additionally check its verdicts.

\FloatBarrier
\begin{figure}[t]
\centering
\includegraphics[width=\linewidth]{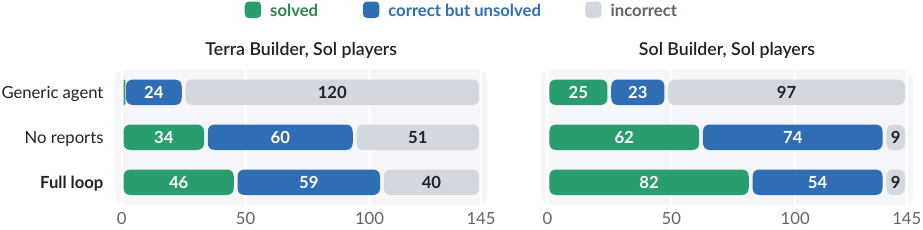}
\caption{\textbf{Task outcomes of all three settings in both experiments.} A task is solved when its environment is correct and the final code policy completes it.}
\label{fig:loop-all}
\end{figure}

\paragraph{Results.}
\Cref{tab:loop-games} gives the outcome of every game in both experiments, and \cref{fig:loop-all} their totals.
With the Terra Builder, the reports make 11 more tasks correct and add 12 solved tasks, at 20\% more model cost.
Defect repairs make the environments of three games correct, and observation requests let the code policies solve tasks in three more games (\cref{tab:loop-repairs}).
The generic agent builds 25 correct environments and solves one task.
It replaced the original game in three games and left it unplayable in one, and in most other tasks its progress or success checks do not follow the game.

With GPT-6.1 Sol as both the Builder and the players, the full loop and the no-reports setting each produce 136 correct task environments, while the number solved by the code policy rises from 62 to 82.
The additional solutions come from six games where the reports lead to richer observations and clearer control guidance, allowing the code policies to solve more tasks without changing which task environments are correct.

\begin{figure*}[p]
\centering
\includegraphics[width=\linewidth]{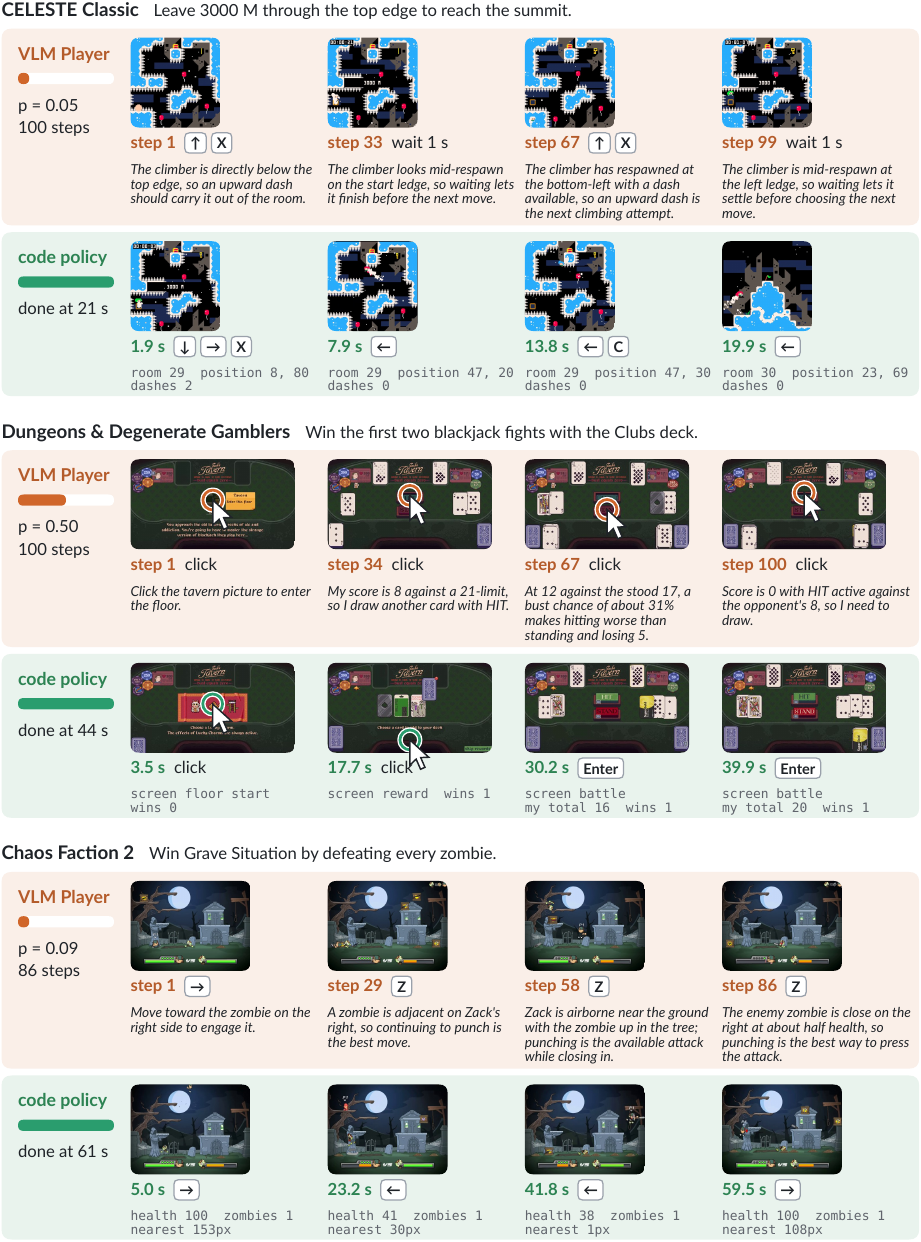}
\caption{\textbf{A \vp and a code policy on the same task.} In each game, the upper row shows four steps of an episode of Claude Haiku 5.5 with the screenshot it saw, its action and its note, and the lower row four moments of a code policy episode that completes the task, each at a key press or click, with the frame, the keys held and fields of the policy's observation. Pointers mark clicks. The bar and $p$ give the progress the \vp reaches.}
\label{fig:players}
\end{figure*}
\begin{figure}[p]
\centering
\includegraphics[width=\linewidth]{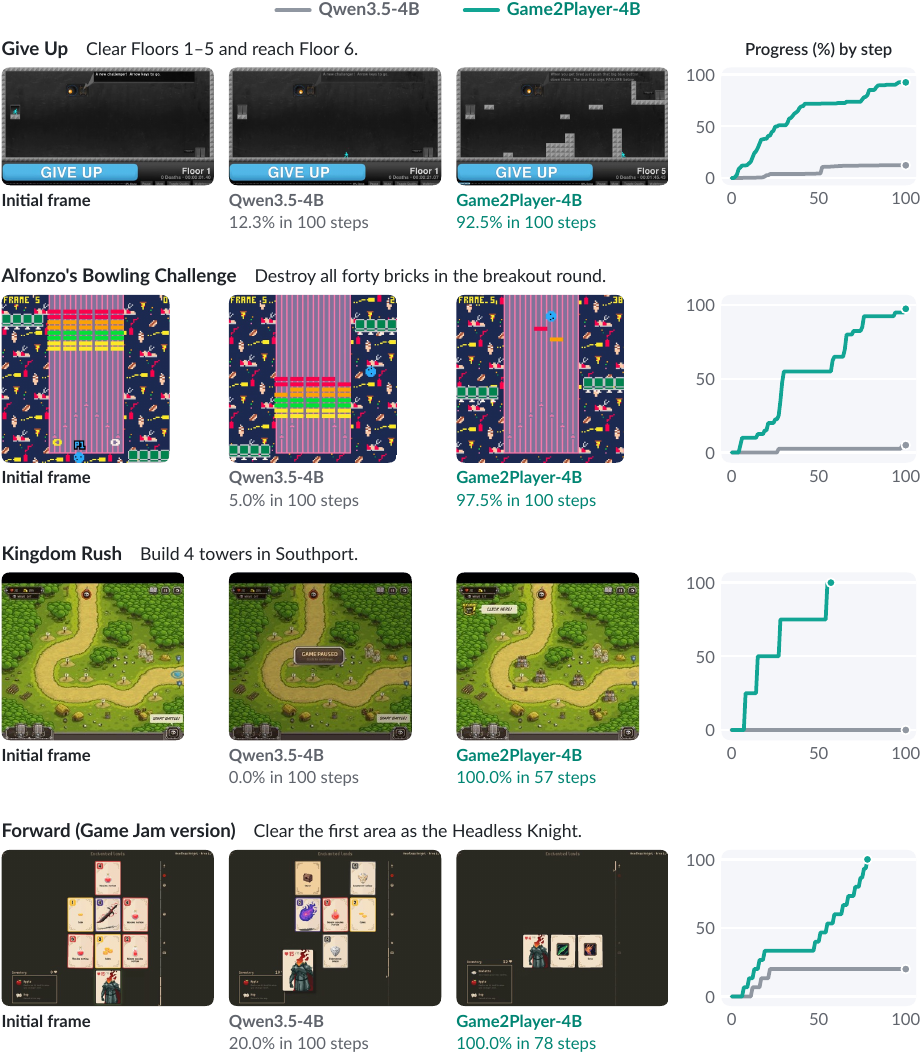}
\caption{\textbf{Base and trained models on held-out \bench tasks}, with thinking on and at most 100 steps. Each row shows the task, the initial frame and the final frames of Qwen3.5-4B and \gfour. The curves show the best progress reached after each step.}
\label{fig:qual}
\end{figure}

\paragraph{Reports and repairs.}
\Cref{tab:loop-repairs} lists every report of the Terra experiment and what the full loop changed.
Eleven games received player reports, and two more had start failures reported by the harness.
The full loop changed eleven environments, with three repairs of defects, four sets of new observations and four clarifications of the control guide, the start or the reported state.
In \emph{There Is No Game}, the \cp found that progress stopped at 0.9 after the letter O had fallen, because the Builder had hooked the game's events before some of their variables existed; after the repair all nine tasks are correct and solved, while without reports all nine remain wrong.
This is typical of the defects that reports catch: the tasks of a game share its instrumentation, so one defect breaks all of them and one repair fixes them all.
Reports of the \vp led to two more repairs, of a crash in \emph{Dungeons \& Degenerate Gamblers} and of an upgrade menu that ran off the screen in \emph{Dome Romantik}.
In \emph{Where the Water Flows} and \emph{Strike Force Heroes}, observation requests made the Builder expose the terrain, the player and the visible enemies, and the code policies solved four tasks in each game, against none without reports.
Of the 568 games in the collection, 298 have complete construction records.
The Builder marked \vp reports as fixed in 23 games and revisited the environment after \cp feedback in 125 games, covering 134 games in total.

\paragraph{Cost.}
\Cref{tab:loop-cost} gives the model cost of each setting per game.
Most of the cost of the full loop and the no-reports setting is the \vp's play, which sends a history of screenshots at every step, and the reports add 20\%, mostly in the Builder's repairs and the players' replays.
\Cref{tab:recording-rates} lists the rates of recording gameplay.
The rates of the \vp{}s come from the 3.4K episodes of our \vp collection and, for GPT-6.1 Sol, from this study, and the cost of writing a code policy comes from the \cp's work on 324 games of the collection.
A \vp costs \$3.2 to \$54.7 per hour of gameplay and takes 3.3 to 19.5 hours of wall-clock time, because the game waits for every decision.
A code policy costs \$0.9 to \$11.6 to write, depending on the model that writes it, and \$0.13 and 1.4 wall-clock hours per hour of gameplay to rerun.
\Cref{fig:cost-models} shows when a code policy pays off.
Written by Claude Opus 5.5 or GPT-6.1 Sol, it is cheaper than GPT-6.1 Sol as a \vp after about 12 minutes of gameplay, than DeepSeek-V4.1-Flash after about an hour and than the cheapest \vp{}s after 2.5 to 4 hours.
Written by DeepSeek-V4.1-Flash, it is cheaper than every \vp within 20 minutes.
In wall-clock time, it is faster than every \vp except Claude Haiku 5.5 within an hour of gameplay, and faster than Haiku after one to three hours.
The steps per hour in \cref{fig:recording} count the decisions of the GPT-6.1 Sol players and the harness steps of the final code policies of the full loop, cut as in \cref{sec:experience}, over the wall-clock time of the same episodes.

\section{Human play}
\label{supp:human}

\paragraph{Protocol.}
Participants play \bench tasks in their own desktop browser, on a web page that serves the same environments as the evaluation with the same seed.
They read the task instruction and the control guide, as the \vp{}s do, and play in real time with keyboard and mouse, for at most 10 minutes per attempt.
An attempt ends when the task's success check fires, when the participant ends it or at the time limit, and a game that allows another try after a loss restarts within the attempt.
The page records the video, the inputs and the state trace in the same format as the other players, and the same evaluator scores every attempt.

\begin{figure}[t]
\centering
\includegraphics[width=\linewidth]{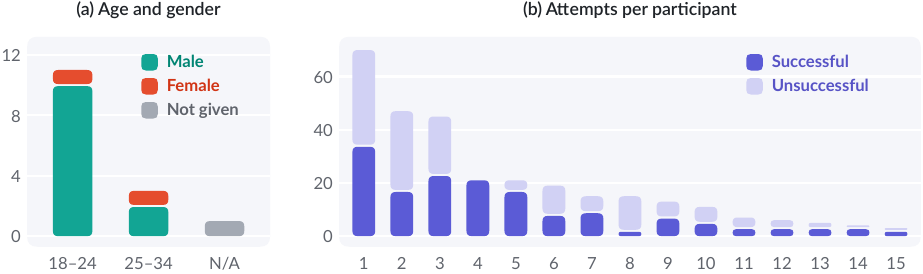}
\caption{\textbf{Participants of the human study.} (a) Age group and gender. (b) Attempts of each participant, split into successful and unsuccessful ones.}
\label{fig:human-participants}
\end{figure}

\paragraph{Participants and coverage.}
The study had 15 participants, recruited through our advertisements in various channels.
Each participant played between 3 and 67 tasks, 19.8 on average, with 302 attempts over all 287 tasks.
Alongside the performance evaluation, human review checked all 287 tasks for playability and scoring consistency.
We checked whether progress reflected advancement toward the stated goal and whether the success conditions matched completion of that goal in the game.
No environment or scoring errors were found.
\Cref{fig:human-participants} summarizes the participants' age and gender and the attempts of each participant.

\paragraph{Consent and privacy.}
Before playing, participants were told on the study page that their gameplay video and their keyboard and mouse input would be recorded for research, and they took part voluntarily and could stop at any time.
The recording covers only the game area of the shared browser tab, without audio, microphone or webcam.
Participants enter a name or nickname, which is stored separately from the recordings, and the age group and gender in \cref{fig:human-participants} are self-reported and shown only in aggregate.

\paragraph{Scoring.}
We score the first attempt of each participant at a task, so that practice on the same task does not count.
A task's progress and success are the means over its scored attempts, and they are then averaged over tasks and games.
Different tasks are played by different participants, so the human row of \cref{tab:players} describes a population of players who meet these games for the first time, not a single expert.
Humans play in real time, while models pause the game at every step, so the human scores are a reference for first-time play rather than an upper bound.
The human scores on \gw in \cref{tab:players} are those of the expert player reported by \gw~\cite{arxiv260407429}, who had studied the game rules and practiced the controls and played each task with at most 100 primitive actions.

\paragraph{Results.}
Human players reach 69.4\% progress and 51.8\% success on \bench.
Their play by difficulty tier is analyzed in \cref{sec:exp-policy} (\cref{fig:human-tiers}).

\FloatBarrier
\section{Qualitative examples}
\label{supp:qual}

\Cref{fig:players} shows how a \vp and a code policy play the same task in three training games, and \cref{fig:qual} compares a base model with \gfour on four held-out tasks from different genres.

More gameplay videos are on the project website at \url{https://game2player.github.io}.

\section{Broader impact}
\label{supp:impact}

The games in our collection belong to their creators.
Only a small share of them come with open-source licenses, and the others are freely playable on public web portals but are not licensed for redistribution.
We use them only for research and do not distribute the games, their assets or the instrumented environments, and all third-party games and assets remain subject to their owners' rights and licenses.
We release the source code of \sys, the trained models and the task specifications.
Other materials remain under review pending copyright and redistribution considerations.
Agents that learn to play games from screenshots could also be used to automate play in online games whose terms of service forbid bots, and our collection contains only single-player games that run offline.

%% file: tables/supp_report_types.tex
\begin{table}[t]
\centering\compacttable[7.8]
\begin{tabularx}{\linewidth}{@{}lX@{}}
\toprule
\rowcolor{tblheadbg}\tblhead{Source} & \tblhead{Allowed content} \\
\midrule
\vp, defect & End of an episode with a report, based on the screenshots, that the game is broken, merged across episodes of a game. \\
\cp, defect & Crash, start routine that misses its situation, key with no effect, progress that never moves, success that fires too early. \\
\cp, observation & A field the code policy needs, its meaning and where it is visible on screen. \\
Not allowed & Hidden information, shortcuts to success, easier tasks. \\
\bottomrule
\end{tabularx}
\par
\caption{\textbf{Reports that return to the Builder.} A defect report describes something that does not work, and an observation request asks for a field of the player observation.}
\label{tab:report-types}
\end{table}

%% file: tables/supp_policy_authors.tex
\begin{table}[h]
\centering\compacttable[7.8]
\setlength{\tabcolsep}{8pt}
\begin{tabular}{lr}
\toprule
\rowcolor{tblheadbg}\tblhead{\cp} & \tblhead{Games} \\
\midrule
GPT-6.1 Sol & 21 \\
Claude Opus 5.5 & 11 \\
Claude Sonnet 5.5 & 5 \\
GPT-6 Luna & 3 \\
DeepSeek-V4.1-Flash & 3 \\
\midrule
No accepted code policy & 7 \\
\bottomrule
\end{tabular}
\par
\caption{\textbf{Coding agents that wrote the code policies of the 50 \bench games.}}
\label{tab:policy-authors}
\end{table}

%% file: figures/float_policy.tex
\begin{figure}[t]
\centering
\includegraphics[width=\colfigwidth]{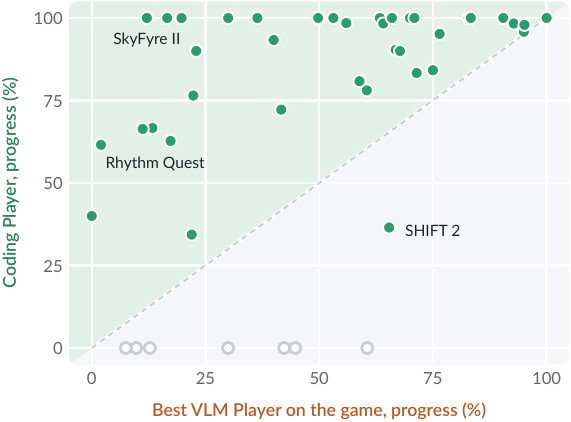}
\caption{\textbf{Code policies against \vp{}s, game by game.} Each dot is one of the 50 \bench games. Dots above the diagonal are games on which the \cp reaches higher progress than all four compared \vp{}s, Qwen3.8-Flash, DeepSeek-V4.1-Flash, GLM-5.3-Flash and Gemini-3.8-Flash.}
\label{fig:policy}
\end{figure}

%% file: tables/supp_policy_per_game.tex
\begin{table*}[tp]
\centering\compacttable[7.2]
\setlength{\tabcolsep}{2.5pt}
\begin{minipage}[t]{0.49\linewidth}\vspace{0pt}
\centering
\begin{tabularx}{\linewidth}{@{}>{\raggedright\arraybackslash}Xrrrr@{}}
\toprule
& \multicolumn{2}{c}{\tblhead{\cp}} & \multicolumn{2}{c@{}}{\tblhead{Best \vp}} \\
\cmidrule(lr){2-3}\cmidrule(l){4-5}
\rowcolor{tblheadbg}\tblhead{Game} & \tblhead{Prog.} & \tblhead{Succ.} & \tblhead{Prog.} & \tblhead{Succ.} \\
\midrule
MOBS, INC & 100.0 & 100.0 & 90.5 & 80.0 \\
Gun Knight & 62.7 & 40.0 & 17.3 & 0.0 \\
Kingdom Rush & 98.3 & 83.3 & 92.8 & 66.7 \\
Zombotron Re-Boot & 0.0 & 0.0 & 12.8 & 11.1 \\
Tiny Kingdom & 76.5 & 50.0 & 22.3 & 0.0 \\
Raide & 40.0 & 28.6 & 0.0 & 0.0 \\
Celeste Classic 2 & 0.0 & 0.0 & 9.8 & 0.0 \\
Chess by Sparsh & 100.0 & 100.0 & 63.3 & 50.0 \\
Memory Match & 100.0 & 100.0 & 100.0 & 100.0 \\
Simon Sequence & 100.0 & 100.0 & 100.0 & 100.0 \\
SHIFT 2 & 36.5 & 28.6 & 65.4 & 42.9 \\
NOM NOM - Great Autumn Jam & 93.3 & 80.0 & 40.0 & 20.0 \\
HIT8OX & 100.0 & 100.0 & 83.3 & 83.3 \\
Pigeon Ascent & 95.8 & 83.3 & 95.0 & 83.3 \\
A Dance of Fire and Ice & 98.3 & 75.0 & 64.1 & 37.5 \\
Auto Necrochess & 78.1 & 40.0 & 60.5 & 40.0 \\
Project: Intellisphere & 72.2 & 33.3 & 41.7 & 0.0 \\
Bonnie's Bakery & 100.0 & 100.0 & 50.0 & 50.0 \\
Frogfall & 100.0 & 100.0 & 66.0 & 0.0 \\
SPLOOB & 100.0 & 100.0 & 30.0 & 20.0 \\
Mindustry Classic & 0.0 & 0.0 & 42.3 & 33.3 \\
A Small World Cup & 83.3 & 50.0 & 71.4 & 50.0 \\
Feed The Void & 100.0 & 100.0 & 69.9 & 50.0 \\
Two-Timin' Towers & 0.0 & 0.0 & 44.8 & 25.0 \\
Petal Quest & 0.0 & 0.0 & 60.6 & 25.0 \\
\bottomrule
\end{tabularx}
\end{minipage}
\hfill
\begin{minipage}[t]{0.49\linewidth}\vspace{0pt}
\centering
\begin{tabularx}{\linewidth}{@{}>{\raggedright\arraybackslash}Xrrrr@{}}
\toprule
& \multicolumn{2}{c}{\tblhead{\cp}} & \multicolumn{2}{c@{}}{\tblhead{Best \vp}} \\
\cmidrule(lr){2-3}\cmidrule(l){4-5}
\rowcolor{tblheadbg}\tblhead{Game} & \tblhead{Prog.} & \tblhead{Succ.} & \tblhead{Prog.} & \tblhead{Succ.} \\
\midrule
Alfonzo's Bowling Challenge & 100.0 & 100.0 & 82.9 & 71.4 \\
Guardian Sphere & 0.0 & 0.0 & 30.0 & 0.0 \\
Packed to the Gills & 100.0 & 100.0 & 36.4 & 0.0 \\
Handshakes & 66.7 & 50.0 & 13.3 & 0.0 \\
Sulka & 80.8 & 60.0 & 58.9 & 20.0 \\
Forward (Game Jam version) & 95.2 & 83.3 & 76.5 & 16.7 \\
Conservation of Bass & 100.0 & 100.0 & 49.8 & 0.0 \\
Snug & 98.5 & 71.4 & 56.0 & 28.6 \\
Dusk Child & 90.0 & 85.7 & 23.0 & 0.0 \\
10-103 & 100.0 & 100.0 & 70.9 & 40.0 \\
William and Sly & 0.0 & 0.0 & 7.5 & 0.0 \\
My Own Hero Party & 84.2 & 66.7 & 75.0 & 50.0 \\
Family Reunion: The Appetizer & 100.0 & 100.0 & 100.0 & 100.0 \\
Last Seen Online & 100.0 & 100.0 & 53.1 & 25.0 \\
About Time! & 100.0 & 100.0 & 97.2 & 83.3 \\
Human Construction Site & 100.0 & 100.0 & 83.3 & 66.7 \\
Pico Puzzle League & 100.0 & 100.0 & 100.0 & 100.0 \\
Mechanical Commando 2: Burning Skies & 66.4 & 40.0 & 11.2 & 0.0 \\
Give Up & 34.4 & 12.5 & 22.0 & 0.0 \\
SkyFyre II & 100.0 & 100.0 & 12.1 & 0.0 \\
Offroaders 2 & 100.0 & 80.0 & 19.7 & 0.0 \\
Fair Squares & 90.4 & 71.4 & 66.8 & 42.9 \\
Bright Lancer & 90.0 & 50.0 & 67.8 & 50.0 \\
Debbie's Diner Derby & 100.0 & 100.0 & 16.6 & 0.0 \\
Rhythm Quest (Demo) & 61.6 & 55.6 & 2.0 & 0.0 \\
\bottomrule
\end{tabularx}
\end{minipage}
\par
\caption{\textbf{Per-game \bench results of the \cp and the best of four \vp{}s.} Progress and success in \%. For each game, the VLM column uses the model with the highest mean progress among Qwen3.8-Flash, DeepSeek-V4.1-Flash, GLM-5.3-Flash and Gemini-3.8-Flash.}
\label{tab:policy-per-game}
\end{table*}

%% file: tables/supp_data_sources.tex
\begin{table*}[tp]
\centering\compacttable[7.8]
\begin{tabularx}{\linewidth}{@{}l*{9}{>{\raggedleft\arraybackslash}X}@{}}
\toprule
\rowcolor{tblheadbg}\tblhead{Source} & \multicolumn{2}{c}{\tblhead{Trajectories}} & \tblhead{Hours} & \tblhead{\shortstack{Eligible\\steps}} & \multicolumn{2}{c}{\tblhead{Training examples}} & \tblhead{\shortstack{Note\\only}} & \tblhead{\shortstack{With\\thinking}} & \tblhead{Neither} \\
\cmidrule(lr){2-3}\cmidrule(lr){6-7}
\rowcolor{tblheadbg} & \tblhead{Count} & \tblhead{\%} & & & \tblhead{Count} & \tblhead{\%} & & & \\
\midrule
\tblbold{\cp} & 12.9K & 81.8 & 217.69 & 1,835.0K & 1,372.1K & 87.7 & 0 & 0 & 1,372.1K \\
\addlinespace[3pt]
\tblbold{\vp} & 2.9K & 18.2 & 88.89 & 219.1K & 191.5K & 12.3 & 169.1K & 22.5K & 0 \\
\quad Claude Haiku 5.5 & 2.2K & 14.3 & 72.53 & 180.0K & 157.1K & 10.0 & 157.1K & 0 & 0 \\
\quad DeepSeek-V4.1-Flash & 0.3K & 1.8 & 5.09 & 15.5K & 14.4K & 0.9 & $<$0.1K & 14.3K & 0 \\
\quad GPT-6 Luna & 0.2K & 1.3 & 7.00 & 14.3K & 11.8K & 0.8 & 11.8K & 0 & 0 \\
\quad MiniMax-M3.1 & 0.1K & 0.8 & 4.27 & 9.4K & 8.2K & 0.5 & 0.1K & 8.1K & 0 \\
\midrule
Total & 15.7K & 100.0 & 306.58 & 2,054.1K & 1,563.6K & 100.0 & 169.1K & 22.5K & 1,372.1K \\
\bottomrule
\end{tabularx}
\par
\caption{\textbf{Recorded gameplay and training examples by source.} Percentages are shares of the full corpus. The four VLM sources appear below their subtotal. Gameplay hours exclude inference pauses, and training examples include resampled copies. The last three columns partition the training examples by the note and thinking at the target step.}
\label{tab:data-sources}
\end{table*}

%% file: tables/supp_training_outcomes.tex
\begin{table}[t]
\centering\compacttable[7.8]
\setlength{\tabcolsep}{8pt}
\begin{tabular}{lrr}
\toprule
\rowcolor{tblheadbg}\tblhead{Source} & \tblhead{Progress} & \tblhead{Success} \\
\midrule
\cp & 97.9 & 96.5 \\
\vp & 39.0 & 21.8 \\
\bottomrule
\end{tabular}
\par
\caption{\textbf{Outcomes of retained training trajectories.} Mean final progress and task success in \%, averaged over recordings. Code policy recordings come from tasks completed by a policy during collection.}
\label{tab:training-outcomes}
\end{table}

%% file: tables/supp_preprocessing.tex
\begin{table}[tp]
\centering\compacttable[7.8]
\setlength{\tabcolsep}{8pt}
\begin{tabular}{lrrr}
\toprule
\rowcolor{tblheadbg}\tblhead{Stage} & \tblhead{Actions} & \tblhead{Waits} & \tblhead{Total} \\
\midrule
Segmented candidates & 1,426.6K & 661.8K & 2,088.5K \\
Invalid actions & $-$13.6K & 0 & $-$13.6K \\
Zero-weight steps & $-$1.2K & $-$19.6K & $-$20.8K \\
Not sampled & $-$379.1K & $-$243.5K & $-$622.6K \\
Repeated uses & $+$132.2K & 0 & $+$132.2K \\
\tblbold{Training examples} & \tblbold{1,164.9K} & \tblbold{398.8K} & \tblbold{1,563.6K} \\
\bottomrule
\end{tabular}
\par
\caption{\textbf{How preprocessing changes the sample count.} Actions and waits are counted separately. Negative rows subtract steps, and the positive row adds repeated uses. Waits with positive weights below one may be omitted during sampling, in addition to waits assigned zero weight.}
\label{tab:preprocessing-counts}
\end{table}

%% file: tables/supp_weighting.tex
\begin{table}[t]
\centering\compacttable[7.8]
\begin{tabularx}{\linewidth}{@{}>{\raggedright\arraybackslash}p{0.2\linewidth}>{\raggedright\arraybackslash}p{0.27\linewidth}>{\raggedright\arraybackslash}X@{}}
\toprule
\rowcolor{tblheadbg}\tblhead{Stage} & \tblhead{Weight} & \tblhead{Application} \\
\midrule
Progress credit & $\geq3,2,1.5$ & Latest three actions ending within $[e-5,e+0.15]$\,s of a progress event at $e$. \\
Success credit & $\geq3,2,1.5$ & Final three eligible actions of a successful episode. \\
Failure suffix & $\times0$ & Up to three actions in the final 5\,s without later progress. \\
Timeout suffix & $\times0.25$ & Up to five actions in the final 10\,s without later progress. \\
Flat repetition & $\times0.35$ & From the third repeated action, if progress does not improve within 5\,s. \\
Game balancing & $B_j$ & Allocate by action type, then in proportion to weight, at most 3 uses per row. \\
Waits & $\min(w_{\text{next}},1)\,0.7^{k}$ & $w_{\text{next}}$ is the next action's weight and $k$ the position in a run of waits. \\
\bottomrule
\end{tabularx}
\par
\caption{\textbf{Sampling weights.} Weights become resampling multiplicities through \cref{eq:pwsft}.}
\label{tab:weighting}
\end{table}

%% file: tables/supp_runs.tex
\begin{minipage}[t]{0.57\linewidth}\vspace{0pt}
\centering\compacttable[7.2]
\begin{tabularx}{\linewidth}{@{}l*{4}{>{\raggedleft\arraybackslash}X}@{}}
\toprule
& \multicolumn{2}{c}{\tblhead{\bench}} & \multicolumn{2}{c}{\tblhead{\gw}} \\
\cmidrule(lr){2-3}\cmidrule(l){4-5}
\rowcolor{tblheadbg}\tblhead{Model} & \tblhead{Progress} & \tblhead{Success} & \tblhead{Progress} & \tblhead{Success} \\
\midrule
\rowcolor{tbloursbg}\gfour & 24.1 $\pm$ 1.6 & 9.9 $\pm$ 2.0 & 33.8 $\pm$ 1.4 & 12.8 $\pm$ 0.9 \\
\rowcolor{tbloursbg}\gnine & 26.4 $\pm$ 2.4 & 12.2 $\pm$ 1.9 & 35.9 $\pm$ 2.5 & 14.6 $\pm$ 1.3 \\
\bottomrule
\end{tabularx}
\par
\captionof{table}{\textbf{Variation across four runs} with thinking off, from seeds 42, 43, 44 and 67. Progress and success in \%, as mean $\pm$ sample standard deviation. The main text reports the run from seed 67.}
\label{tab:repeat}

\vspace{10pt}
\begin{tabularx}{\linewidth}{@{}>{\raggedright\arraybackslash}Xrr*{4}{>{\raggedleft\arraybackslash}p{4.4em}}@{}}
\toprule
& & & \multicolumn{2}{c}{\tblhead{DeepSeek-V4.1-Flash}} & \multicolumn{2}{c}{\tblhead{Qwen3.8-Flash}} \\
\cmidrule(lr){4-5}\cmidrule(l){6-7}
\rowcolor{tblheadbg}\tblhead{Setting} & \tblhead{Text} & \tblhead{Images} & \tblhead{Prog.} & \tblhead{Succ.} & \tblhead{Prog.} & \tblhead{Succ.} \\
\midrule
Full & all & 6 & 39.5 & 23.3 & 45.1 & 24.7 \\
Short text & 6 & 6 & 42.4 & 22.0 & 38.9 & 19.5 \\
Fewer images & 50 & 2 & 46.0 & 26.5 & 44.0 & 25.1 \\
Short both & 6 & 2 & 42.4 & 23.7 & 38.3 & 20.6 \\
\bottomrule
\end{tabularx}
\par
\captionof{table}{\textbf{History length on \bench.} Progress and success in \%. The current screenshot is always included.}
\label{tab:history}
\end{minipage}\hfill
\begin{minipage}[t]{0.40\linewidth}\vspace{0pt}
\centering\compacttable[7.2]
\begin{tabularx}{\linewidth}{@{}l*{4}{>{\raggedleft\arraybackslash}X}@{}}
\toprule
& \multicolumn{2}{c}{\tblhead{Progress}} & \multicolumn{2}{c}{\tblhead{Success}} \\
\cmidrule(lr){2-3}\cmidrule(l){4-5}
\rowcolor{tblheadbg}\tblhead{Model} & \tblhead{Ours} & \tblhead{GW} & \tblhead{Ours} & \tblhead{GW} \\
\midrule
Claude Opus 5.5 & 65.5 & 64.3 & 47.6 & 42.9 \\
GPT-6.1 Sol & 64.3 & 71.8 & 44.1 & 54.1 \\
GLM-5.3-Flash & 48.6 & 34.5 & 25.9 & 14.7 \\
Qwen3.8-27B & 47.7 & 49.7 & 25.3 & 27.6 \\
Qwen3.8-Flash & 44.4 & 52.0 & 22.4 & 26.5 \\
Qwen3.5-35B-A3B & 34.3 & 34.7 & 15.3 & 14.1 \\
\bottomrule
\end{tabularx}
\par
\captionof{table}{\textbf{Harness comparison on \gw.} GW denotes the generalist harness.}
\label{tab:harness-comparison}
\end{minipage}

%% file: tables/tab_codev_games.tex
\compacttable[7.2]
\begin{tabularx}{\linewidth}{@{}>{\raggedright\arraybackslash}Xlr*{4}{>{\hspace{4pt}}rrr}@{}}
\toprule
& & & \multicolumn{6}{c}{\tblhead{Terra Builder, Sol players}} & \multicolumn{6}{c}{\tblhead{Sol Builder, Sol players}} \\
\cmidrule(lr){4-9}\cmidrule(l){10-15}
& & & \multicolumn{3}{c}{\tblhead{Correct}} & \multicolumn{3}{c}{\tblhead{Solved}} & \multicolumn{3}{c}{\tblhead{Correct}} & \multicolumn{3}{c}{\tblhead{Solved}} \\
\cmidrule(lr){4-6}\cmidrule(lr){7-9}\cmidrule(lr){10-12}\cmidrule(l){13-15}
\rowcolor{tblheadbg}\tblhead{Game} & \tblhead{Engine} & \tblhead{Tasks} & \tblhead{Gen} & \tblhead{NoR} & \tblhead{Full} & \tblhead{Gen} & \tblhead{NoR} & \tblhead{Full} & \tblhead{Gen} & \tblhead{NoR} & \tblhead{Full} & \tblhead{Gen} & \tblhead{NoR} & \tblhead{Full} \\
\midrule
Drift Hunters & Unity WebGL & 7 & 0 & 0 & 0 & 0 & 0 & 0 & 0 & 0 & 0 & 0 & 0 & 0 \\
Backpack Hero & Unity WebGL & 4 & 0 & 4 & 4 & 0 & 0 & 0 & 1 & 4 & 4 & 1 & 3 & 3 \\
Peggy's Post & Unity WebGL & 5 & 0 & 1 & 1 & 0 & 0 & 0 & 0 & 4 & 4 & 0 & 2 & 2 \\
Rain on Your Parade & Unity WebGL & 8 & 0 & 7 & 7 & 0 & 6 & 5 & 0 & 8 & 8 & 0 & 6 & 6 \\
Stick to the Plan & Unity WebGL & 5 & 0 & 4 & 4 & 0 & 4 & 4 & 0 & 5 & 5 & 0 & 3 & 3 \\
Where the Water Flows & Godot & 7 & 0 & 7 & 7 & 0 & 0 & 4 & 0 & 7 & 7 & 0 & 0 & 0 \\
Dungeons \& Degenerate Gamblers & Godot & 7 & 0 & 6 & 7 & 0 & 0 & 0 & 7 & 7 & 7 & 5 & 0 & 5 \\
HELP! NO BRAKE & Godot & 10 & 0 & 10 & 10 & 0 & 8 & 8 & 0 & 10 & 10 & 0 & 8 & 9 \\
Dome Romantik & Godot & 6 & 0 & 5 & 6 & 0 & 4 & 0 & 2 & 6 & 6 & 2 & 4 & 5 \\
SHRUBNAUT & GameMaker & 6 & 6 & 6 & 6 & 1 & 0 & 0 & 6 & 6 & 6 & 1 & 2 & 2 \\
Dead Estate & GameMaker & 9 & 6 & 6 & 6 & 0 & 0 & 0 & 9 & 9 & 9 & 3 & 4 & 4 \\
PortaBoy+ & GameMaker & 5 & 5 & 5 & 5 & 0 & 5 & 5 & 5 & 5 & 5 & 5 & 5 & 5 \\
Hardware Tycoon & Construct & 8 & 0 & 0 & 0 & 0 & 0 & 0 & 0 & 7 & 7 & 0 & 7 & 7 \\
SUPERHOTline Miami & Construct & 9 & 8 & 0 & 0 & 0 & 0 & 0 & 9 & 9 & 9 & 2 & 2 & 2 \\
There Is No Game & Construct & 9 & 0 & 0 & 9 & 0 & 0 & 9 & 8 & 9 & 9 & 6 & 9 & 9 \\
CELESTE Classic & PICO-8 & 10 & 0 & 10 & 10 & 0 & 4 & 5 & 0 & 10 & 10 & 0 & 3 & 6 \\
Slipways & PICO-8 & 9 & 0 & 9 & 9 & 0 & 0 & 0 & 1 & 9 & 9 & 0 & 0 & 6 \\
Strike Force Heroes & Flash & 5 & 0 & 5 & 5 & 0 & 0 & 4 & 0 & 5 & 5 & 0 & 0 & 4 \\
Chronotron & Flash & 8 & 0 & 1 & 1 & 0 & 1 & 0 & 0 & 8 & 8 & 0 & 0 & 0 \\
Tiny Truck Racing & Defold & 8 & 0 & 8 & 8 & 0 & 2 & 2 & 0 & 8 & 8 & 0 & 4 & 4 \\
\midrule
Total & 7 engines & 145 & 25 & 94 & 105 & 1 & 34 & 46 & 48 & 136 & 136 & 25 & 62 & 82 \\
\bottomrule
\end{tabularx}
\par
\captionof{table}{\textbf{Per-game results of the co-development study.} Tasks with a correct environment and solved tasks in both experiments, for the generic agent (Gen), the no-reports setting (NoR) and the full loop (Full).}
\label{tab:loop-games}

%% file: tables/tab_codev_repairs.tex
\compacttable[7.2]
\begin{tabularx}{\linewidth}{@{}l>{\raggedright\arraybackslash}X>{\raggedright\arraybackslash}Xrr@{}}
\toprule
\rowcolor{tblheadbg}\tblhead{Game} & \tblhead{Report} & \tblhead{Change in the full loop} & \tblhead{Correct} & \tblhead{Solved} \\
\midrule
There Is No Game & \cp: progress stays at 0.9 after the letter O falls & Hooks each game variable when it appears, so later story events are recorded & 0/9 & 0/9 \\
Dungeons \& Degenerate Gamblers & \vp: the table stays empty after the Gambler is chosen; \cp: cards and buttons cannot be read & Cancels pending room animations before freeing the cards; adds cards, buttons and battle state & 6/7 & 0/0 \\
Dome Romantik & \vp: the upgrade list runs below the screen & Keeps five upgrades on screen and scrolls the list with the selection & 5/6 & 4/0 \\
Where the Water Flows & \cp: terrain and objects cannot be read & Adds terrain, platforms, mechanisms and movable objects & 7/7 & 0/4 \\
Strike Force Heroes & \cp: the player, soldiers and flags cannot be read & Adds the player, visible units, flags and pickups & 5/5 & 0/4 \\
CELESTE Classic & \cp: moving objects cannot be read & Adds moving platforms, balloons, springs and breakable blocks & 10/10 & 4/5 \\
HELP! NO BRAKE & \cp: heading and obstacles cannot be read & Adds heading, motion, terrain and saw blades & 10/10 & 8/8 \\
Chronotron & \vp: the timer stops in the pod; \cp: surfaces and actors cannot be read & Guide: all past selves must return to the pod; adds robots and switches & 1/1 & 1/0 \\
Slipways & \vp: the score overlay stays open; \cp: the map and trades cannot be read & Guide: a panel closes after its animation; adds planets, links and menus & 9/9 & 0/0 \\
Dead Estate & \cp: a white screen after the shop & Reports a stalled game loop as not actionable & 6/6 & 0/0 \\
SUPERHOTline Miami & \cp: one task fails to start & None & 0/0 & 0/0 \\
Rain on Your Parade & Harness: the game starts too slowly & Speeds up the opening cutscene while input is blocked & 7/7 & 6/5 \\
Tiny Truck Racing & Harness: the game starts too slowly & None & 8/8 & 2/2 \\
\bottomrule
\end{tabularx}
\par
\captionof{table}{\textbf{Reports and repairs with the GPT-6.1 Terra Builder.} Harness reports are start failures that the harness detects by itself. Correct and Solved give the number of tasks without reports / with the full loop.}
\label{tab:loop-repairs}

%% file: tables/tab_codev_cost.tex
\compacttable[7.8]
\setlength{\tabcolsep}{7pt}
\renewcommand{\tblsection}[2]{\multicolumn{#1}{l}{\cellcolor{tblgroupbg}\tblhead{#2}}\\}
\begin{tabularx}{\linewidth}{>{\raggedright\arraybackslash}X*{4}{r}}
\toprule
\rowcolor{tblheadbg}\tblhead{Setting} & \tblhead{Builder} & \tblhead{\vp} & \tblhead{\cp} & \tblhead{Total} \\
\midrule
\tblsection{5}{Terra Builder, Sol players}
Generic agent & 0.12 & -- & 0.17 & 0.29 \\
No reports & 1.41 & 9.59 & 6.66 & 17.66 \\
Full loop & 2.96 & 12.23 & 6.06 & 21.25 \\
\midrule
\tblsection{5}{Sol Builder, Sol players}
Generic agent & 1.09 & -- & 3.36 & 4.45 \\
No reports & 3.67 & 10.73 & 7.68 & 22.08 \\
Full loop & 4.95 & 11.05 & 10.23 & 26.23 \\
\bottomrule
\end{tabularx}
\par
\captionof{table}{\textbf{Model cost of the co-development study,} in dollars per game. The generic agent has no \vp.}
\label{tab:loop-cost}

%% file: tables/tab_recording_rates.tex
\compacttable[7.8]
\begin{tabularx}{\linewidth}{@{}ll*{4}{>{\raggedleft\arraybackslash}X}r*{2}{>{\raggedleft\arraybackslash}X}@{}}
\toprule
& & \multicolumn{3}{c}{\tblhead{Tokens per hour (M)}} & \multicolumn{4}{c}{\tblhead{Per hour of gameplay}} \\
\cmidrule(lr){3-5}\cmidrule(l){6-9}
\rowcolor{tblheadbg}\tblhead{\vp} & \tblhead{Price} & \tblhead{Input} & \tblhead{Cached} & \tblhead{Output} & \tblhead{Model \$} & \tblhead{Wall-clock h} & \tblhead{CPU \$} & \tblhead{Total \$} \\
\midrule
Claude Haiku 5.5 & 0.1/0.01/0.5 & 31.0 & 19\% & 0.65 & 2.91 & 3.3 & 0.30 & 3.21 \\
GPT-6 Luna & 0.1/0.01/0.5 & 55.9 & 55\% & 0.41 & 3.04 & 11.5 & 1.04 & 4.08 \\
MiniMax-M3.1 & 0.15/0.015/0.6 & 24.3 & 17\% & 0.93 & 3.65 & 7.1 & 0.64 & 4.29 \\
DeepSeek-V4.1-Flash & 0.3/0.006/1.2 & 23.6 & 45\% & 4.68 & 9.56 & 19.5 & 1.76 & 11.31 \\
GPT-6.1 Sol & 2/0.1/10 & 34.9 & 30\% & 0.43 & 54.06 & 7.6 & 0.68 & 54.74 \\
\bottomrule
\end{tabularx}

\vspace{6pt}
{\setlength{\tabcolsep}{8pt}%
\begin{tabular}{lrr}
\toprule
\rowcolor{tblheadbg}\tblhead{Writer of the code policy} & \tblhead{\$ per policy} & \tblhead{Hours per policy} \\
\midrule
Claude Opus 5.5 & 11.56 & 2.2 \\
GPT-6.1 Sol & 10.80 & 3.4 \\
DeepSeek-V4.1-Flash & 0.89 & 4.8 \\
\bottomrule
\end{tabular}}
\par
\captionof{table}{\textbf{Rates of recording gameplay.} Prices are list prices in dollars per million input, cached input and output tokens, and each browser runs on two vCPUs at \$0.045 per vCPU-hour. Wall-clock time is the real time that recording takes, including the pauses while a \vp decides. Reruns of a code policy call no model and cost \$0.13 and 1.4 wall-clock hours per hour of gameplay.}
\label{tab:recording-rates}

%% file: main.bbl
\begin{thebibliography}{80}
\providecommand{\natexlab}[1]{#1}
\providecommand{\url}[1]{\texttt{#1}}
\expandafter\ifx\csname urlstyle\endcsname\relax
  \providecommand{\doi}[1]{doi: #1}\else
  \providecommand{\doi}{doi: \begingroup \urlstyle{rm}\Url}\fi

\bibitem[Ahn et~al.(2025)Ahn, Kim, Yun, Son, Park, Cho, and
  Kim]{arxiv250901052}
Jaewoo Ahn, Junseo Kim, Heeseung Yun, Jaehyeon Son, Dongmin Park, Jaewoong Cho,
  and Gunhee Kim.
\newblock Flashadventure: A benchmark for gui agents solving full story arcs in
  diverse adventure games.
\newblock In \emph{Proceedings of the 2025 Conference on Empirical Methods in
  Natural Language Processing}, pages 23365--23395, 2025.

\bibitem[Baker et~al.(2022)Baker, Akkaya, Zhokov, Huizinga, Tang, Ecoffet,
  Houghton, Sampedro, and Clune]{baker2022video}
Bowen Baker, Ilge Akkaya, Peter Zhokov, Joost Huizinga, Jie Tang, Adrien
  Ecoffet, Brandon Houghton, Raul Sampedro, and Jeff Clune.
\newblock Video pretraining (vpt): Learning to act by watching unlabeled online
  videos.
\newblock \emph{Advances in Neural Information Processing Systems},
  35:\penalty0 24639--24654, 2022.

\bibitem[Bolton et~al.(2025)Bolton, Lerchner, Cordell, Moufarek, Bolt,
  Lampinen, Mitenkova, Hallingstad, Vujatovic, Li, et~al.]{simateam2025sima}
Adrian Bolton, Alexander Lerchner, Alexandra Cordell, Alexandre Moufarek,
  Andrew Bolt, Andrew Lampinen, Anna Mitenkova, Arne~Olav Hallingstad, Bojan
  Vujatovic, Bonnie Li, et~al.
\newblock Sima 2: A generalist embodied agent for virtual worlds.
\newblock \emph{arXiv preprint arXiv:2512.04797}, 2025.

\bibitem[Cao et~al.(2024)Cao, Tang, Zhao, Guo, Liu, Zhang, Liu, Sun, Reid, and
  Liang]{cao2024physgame}
Meng Cao, Haoran Tang, Haoze Zhao, Hangyu Guo, Jiaheng Liu, Ge~Zhang, Ruyang
  Liu, Qiang Sun, Ian Reid, and Xiaodan Liang.
\newblock Physgame: Uncovering physical commonsense violations in gameplay
  videos.
\newblock \emph{arXiv preprint arXiv:2412.01800}, 2024.

\bibitem[Chae et~al.(2026)Chae, Park, and Ritter]{chae2026safeandscala}
Hyungjoo Chae, Jungsoo Park, and Alan Ritter.
\newblock Safe and scalable web agent learning via recreated websites.
\newblock \emph{arXiv preprint arXiv:2603.10505}, 2026.

\bibitem[Chen et~al.(2025{\natexlab{a}})Chen, Zhao, Yang, Ma, Yang, Yao, Wang,
  Bai, Wang, Pan, et~al.]{chen2025era}
Hanyang Chen, Mark Zhao, Rui Yang, Qinwei Ma, Ke~Yang, Jiarui Yao, Kangrui
  Wang, Hao Bai, Zhenhailong Wang, Rui Pan, et~al.
\newblock Era: Transforming vlms into embodied agents via embodied prior
  learning and online reinforcement learning.
\newblock \emph{arXiv preprint arXiv:2510.12693}, 2025{\natexlab{a}}.

\bibitem[Chen et~al.(2025{\natexlab{b}})Chen, Gao, Liu, Huang, Sung, Zhou, Wu,
  and Chang]{arxiv250513426}
Liang Chen, Hongcheng Gao, Tianyu Liu, Zhiqi Huang, Flood Sung, Xinyu Zhou,
  Yuxin Wu, and Baobao Chang.
\newblock G1: Bootstrapping perception and reasoning abilities of
  vision-language model via reinforcement learning.
\newblock \emph{arXiv preprint arXiv:2505.13426}, 2025{\natexlab{b}}.

\bibitem[Chen et~al.(2026)Chen, Zhai, Wang, Yang, Zhang, Wang, and
  Zhang]{chen2026recap}
Yuxi Chen, Haoyu Zhai, Chenkai Wang, Rui Yang, Lingming Zhang, Gang Wang, and
  Huan Zhang.
\newblock {CAPTCHA} solving for native {GUI} agents: Automated reasoning-action
  data generation and self-corrective training.
\newblock In \emph{International Conference on Machine Learning}, 2026.

\bibitem[Choi et~al.(2026)Choi, Jung, Seong, Kim, Kim, Cho, Kim, Park, Yu, and
  Lee]{choi2025d2e}
Suhwan Choi, Jaeyoon Jung, Haebin Seong, Minchan Kim, Minyeong Kim, Yongjun
  Cho, Yoonshik Kim, Yubeen Park, Youngjae Yu, and Yunsung Lee.
\newblock D2e: Scaling vision-action pretraining on desktop data for transfer
  to embodied ai.
\newblock In \emph{International Conference on Learning Representations},
  volume 2026, pages 46207--46236, 2026.

\bibitem[Choudhury and Sodhi(2025)]{choudhury2024better}
Sanjiban Choudhury and Paloma Sodhi.
\newblock Better than your teacher: Llm agents that learn from privileged ai
  feedback.
\newblock In \emph{International Conference on Learning Representations},
  volume 2025, pages 10457--10493, 2025.

\bibitem[Cloos et~al.(2024)Cloos, Jens, Naim, Kuo, Cases, Barbu, and
  Cueva]{arxiv240713729}
Nathan Cloos, Meagan Jens, Michelangelo Naim, Yen-Ling Kuo, Ignacio Cases,
  Andrei Barbu, and Christopher~J Cueva.
\newblock Baba is ai: Break the rules to beat the benchmark.
\newblock \emph{arXiv preprint arXiv:2407.13729}, 2024.

\bibitem[Dennis et~al.(2020)Dennis, Jaques, Vinitsky, Bayen, Russell, Critch,
  and Levine]{arxiv201202096}
Michael Dennis, Natasha Jaques, Eugene Vinitsky, Alexandre Bayen, Stuart
  Russell, Andrew Critch, and Sergey Levine.
\newblock Emergent complexity and zero-shot transfer via unsupervised
  environment design.
\newblock \emph{Advances in neural information processing systems},
  33:\penalty0 13049--13061, 2020.

\bibitem[Faldor et~al.(2025)Faldor, Zhang, Cully, and
  Clune]{faldor2025omniepic}
Maxence Faldor, Jenny Zhang, Antoine Cully, and Jeff Clune.
\newblock {OMNI-EPIC}: Open-endedness via models of human notions of
  interestingness with environments programmed in code.
\newblock In \emph{International Conference on Learning Representations}, 2025.

\bibitem[Foundation(2026)]{arxiv260324621}
ARC Foundation.
\newblock Arc-agi-3: A new challenge for frontier agentic intelligence.
\newblock \emph{arXiv preprint arXiv:2603.24621}, 2026.

\bibitem[Gandhi et~al.(2026)Gandhi, Garg, Goodman, and
  Papailiopoulos]{arxiv260116443}
Kanishk Gandhi, Shivam Garg, Noah~D Goodman, and Dimitris Papailiopoulos.
\newblock Endless terminals: Scaling rl environments for terminal agents.
\newblock \emph{arXiv preprint arXiv:2601.16443}, 2026.

\bibitem[Guss et~al.(2019)Guss, Houghton, Topin, Wang, Codel, Veloso, and
  Salakhutdinov]{guss2019minerl}
William~H Guss, Brandon Houghton, Nicholay Topin, Phillip Wang, Cayden Codel,
  Manuela Veloso, and Ruslan Salakhutdinov.
\newblock Minerl: A large-scale dataset of minecraft demonstrations.
\newblock \emph{arXiv preprint arXiv:1907.13440}, 2019.

\bibitem[Han et~al.(2026)Han, Hu, Qiu, Wu, and He]{han2026vista}
Qiushi Han, Keya Hu, Linlu Qiu, Cathy Wu, and Kaiming He.
\newblock Vista: A visual harness for reasoning in an interactive world.
\newblock \emph{arXiv preprint arXiv:2610.02200}, 2026.

\bibitem[He et~al.(2026)He, Chawla, Souri, Som, and Song]{he2025webstar}
Yifei He, Pranit Chawla, Yaser Souri, Subhojit Som, and Xia Song.
\newblock Webstar: Scalable data synthesis for computer use agents with
  step-level filtering.
\newblock In \emph{Proceedings of the 64th Annual Meeting of the Association
  for Computational Linguistics (Volume 1: Long Papers)}, pages 516--533, 2026.

\bibitem[He et~al.(2025)He, Weilbach, Wojciechowska, Zhang, and
  Wood]{he2025plaicraft}
Yingchen He, Christian~D Weilbach, Martyna~E Wojciechowska, Yuxuan Zhang, and
  Frank Wood.
\newblock Plaicraft: Large-scale time-aligned vision-speech-action dataset for
  embodied ai.
\newblock \emph{arXiv preprint arXiv:2505.12707}, 2025.

\bibitem[Hu et~al.(2026)Hu, Huo, Zhang, Yu, Xing, Stoica, Rosing, Jin, and
  Zhang]{arxiv250515146}
Lanxiang Hu, Mingjia Huo, Yuxuan Zhang, Haoyang Yu, Eric~P Xing, Ion Stoica,
  Tajana Rosing, Haojian Jin, and Hao Zhang.
\newblock lmgame-bench: How good are llms at playing games?
\newblock In \emph{International Conference on Learning Representations},
  volume 2026, pages 81308--81356, 2026.

\bibitem[Hu and Clune(2023)]{hu2023thought}
Shengran Hu and Jeff Clune.
\newblock Thought cloning: Learning to think while acting by imitating human
  thinking.
\newblock \emph{Advances in Neural Information Processing Systems},
  36:\penalty0 44451--44469, 2023.

\bibitem[Jain et~al.(2025)Jain, Singh, Shetty, Zheng, Sen, and
  Stoica]{jain2025r2egym}
Naman Jain, Jaskirat Singh, Manish Shetty, Liang Zheng, Koushik Sen, and Ion
  Stoica.
\newblock R2e-gym: Procedural environments and hybrid verifiers for scaling
  open-weights swe agents.
\newblock \emph{arXiv preprint arXiv:2504.07164}, 2025.

\bibitem[Kurin et~al.(2017)Kurin, Nowozin, Hofmann, Beyer, and
  Leibe]{kurin2017atari}
Vitaly Kurin, Sebastian Nowozin, Katja Hofmann, Lucas Beyer, and Bastian Leibe.
\newblock The atari grand challenge dataset.
\newblock \emph{arXiv preprint arXiv:1705.10998}, 2017.

\bibitem[Levenshtein(1966)]{levenshtein1966binary}
Vladimir~I. Levenshtein.
\newblock Binary codes capable of correcting deletions, insertions, and
  reversals.
\newblock \emph{Soviet Physics Doklady}, 10\penalty0 (8):\penalty0 707--710,
  1966.

\bibitem[Li et~al.(2026)Li, Yang, Qian, Liu, Kim, Wang, Li, Zhang, and
  Ji]{li2026memcompress}
Bingxuan Li, Rui Yang, Cheng Qian, Jiateng Liu, Jeonghwan Kim, Zhenhailong
  Wang, Manling Li, Tong Zhang, and Heng Ji.
\newblock Long-horizon embodied decision-making via multimodal memory
  compression.
\newblock \emph{arXiv preprint arXiv:2608.01456}, 2026.

\bibitem[Li et~al.(2025)Li, Wang, He, Ma, and Liang]{li2025jarvis}
Muyao Li, Zihao Wang, Kaichen He, Xiaojian Ma, and Yitao Liang.
\newblock Jarvis-vla: Post-training large-scale vision language models to play
  visual games with keyboards and mouse.
\newblock In \emph{Findings of the Association for Computational Linguistics:
  ACL 2025}, pages 17878--17899, 2025.

\bibitem[Lin et~al.(2026)Lin, Qian, Liu, Huang, Wang, Huang, Li, Zhang, Hu,
  Zhu, et~al.]{arxiv260609826}
Mingxian Lin, Shengju Qian, Yuqi Liu, Yi-Hua Huang, Yiyu Wang, Wei Huang,
  Yitang Li, Fan Zhang, Zeyu Hu, Lingting Zhu, et~al.
\newblock Omnigamearena: A unified ue5 benchmark for vlm game agents with
  improvement dynamics.
\newblock \emph{arXiv preprint arXiv:2606.09826}, 2026.

\bibitem[Magne et~al.(2026)Magne, Awadalla, Wang, Xu, Belofsky, Hu, Kim,
  Schmidt, Gkioxari, Kautz, et~al.]{arxiv260102427}
Lo{\"\i}c Magne, Anas Awadalla, Guanzhi Wang, Yinzhen Xu, Joshua Belofsky,
  Fengyuan Hu, Joohwan Kim, Ludwig Schmidt, Georgia Gkioxari, Jan Kautz, et~al.
\newblock Nitrogen: An open foundation model for generalist gaming agents.
\newblock \emph{arXiv preprint arXiv:2601.02427}, 2026.

\bibitem[Mun et~al.(2017)Mun, Hongsuck~Seo, Jung, and Han]{mun2016marioqa}
Jonghwan Mun, Paul Hongsuck~Seo, Ilchae Jung, and Bohyung Han.
\newblock Marioqa: Answering questions by watching gameplay videos.
\newblock In \emph{Proceedings of the IEEE International Conference on Computer
  Vision}, pages 2867--2875, 2017.

\bibitem[Ouyang et~al.(2026{\natexlab{a}})Ouyang, Hu, Lin, Ng, and
  Shou]{arxiv260407429}
Mingyu Ouyang, Siyuan Hu, Kevin~Qinghong Lin, Hwee~Tou Ng, and Mike~Zheng Shou.
\newblock Gameworld: Towards standardized and verifiable evaluation of
  multimodal game agents.
\newblock In \emph{European Conference on Computer Vision}, pages 41--59.
  Springer, 2026{\natexlab{a}}.

\bibitem[Ouyang et~al.(2026{\natexlab{b}})Ouyang, Wang, Hu, Yin, Sun, Wang, Pu,
  Cheng, Chen, Shan, et~al.]{ouyang2026player1}
Mingyu Ouyang, Yiran Wang, Siyuan Hu, Xingyilang Yin, Huiqiang Sun, Guangzhi
  Wang, Junfu Pu, Cheng Cheng, Honghao Chen, Ying Shan, et~al.
\newblock Player-1: Learning to play from synthetic experience.
\newblock \emph{arXiv preprint arXiv:2610.20584}, 2026{\natexlab{b}}.

\bibitem[Paglieri et~al.(2025)Paglieri, Cupia{\l}, Coward, Piterbarg,
  Wo{\l}czyk, Khan, Pignatelli, Kuci{\'n}ski, Pinto, Fergus,
  et~al.]{arxiv241113543}
Davide Paglieri, Bart{\l}omiej Cupia{\l}, Samuel Coward, Ulyana Piterbarg,
  Maciej Wo{\l}czyk, Akbir Khan, Eduardo Pignatelli, {\L}ukasz Kuci{\'n}ski,
  Lerrel Pinto, Rob Fergus, et~al.
\newblock Balrog: Benchmarking agentic llm and vlm reasoning on games.
\newblock In \emph{International Conference on Learning Representations},
  volume 2025, pages 96666--96702, 2025.

\bibitem[Park et~al.(2026)Park, Kim, Choi, Kim, Lee, Lee, Park, Lee, Hwang,
  Ahn, et~al.]{arxiv250603610}
Dongmin Park, Minkyu Kim, Beongjun Choi, Junhyuck Kim, Keon Lee, Jonghyun Lee,
  Inkyu Park, ByeongUk Lee, Jaeyoung Hwang, Jaewoo Ahn, et~al.
\newblock Orak: A foundational benchmark for training and evaluating llm agents
  on diverse video games.
\newblock In \emph{International Conference on Learning Representations},
  volume 2026, pages 60684--60744, 2026.

\bibitem[Pearce and Zhu(2022)]{pearce2021counter}
Tim Pearce and Jun Zhu.
\newblock Counter-strike deathmatch with large-scale behavioural cloning.
\newblock In \emph{2022 IEEE Conference on Games (CoG)}, pages 104--111. IEEE,
  2022.

\bibitem[Penaloza et~al.(2026)Penaloza, Vattikonda, Gontier, Lacoste, Charlin,
  and Caccia]{penaloza2026privileged}
Emiliano Penaloza, Dheeraj Vattikonda, Nicolas Gontier, Alexandre Lacoste,
  Laurent Charlin, and Massimo Caccia.
\newblock Privileged information distillation for language models.
\newblock \emph{arXiv preprint arXiv:2602.04942}, 2026.

\bibitem[Qiao et~al.(2026)Qiao, Tang, Hao, and Li]{arxiv260922308}
Boyu Qiao, Zixin Tang, Xiaoshuai Hao, and Wenbo Li.
\newblock Gamereplica: A benchmark for black-box visual game replication by
  vision-language agents.
\newblock \emph{arXiv preprint arXiv:2609.22308}, 2026.

\bibitem[Qin et~al.(2025)Qin, Ye, Fang, Wang, Liang, Tian, Zhang, Li, Li,
  Huang, et~al.]{qin2025ui}
Yujia Qin, Yining Ye, Junjie Fang, Haoming Wang, Shihao Liang, Shizuo Tian,
  Junda Zhang, Jiahao Li, Yunxin Li, Shijue Huang, et~al.
\newblock Ui-tars: Pioneering automated gui interaction with native agents.
\newblock \emph{arXiv preprint arXiv:2501.12326}, 2025.

\bibitem[{Qwen Team}(2026)]{qwen35}
{Qwen Team}.
\newblock {Qwen3.5}: Towards native multimodal agents, February 2026.
\newblock URL \url{https://qwen.ai/blog?id=qwen3.5}.

\bibitem[Raad et~al.(2024)Raad, Ahuja, Barros, Besse, Bolt, Bolton, Brownfield,
  Buttimore, Cant, Chakera, et~al.]{arxiv240410179}
Maria~Abi Raad, Arun Ahuja, Catarina Barros, Frederic Besse, Andrew Bolt,
  Adrian Bolton, Bethanie Brownfield, Gavin Buttimore, Max Cant, Sarah Chakera,
  et~al.
\newblock Scaling instructable agents across many simulated worlds.
\newblock \emph{arXiv preprint arXiv:2404.10179}, 2024.

\bibitem[Rubner et~al.(2000)Rubner, Tomasi, and Guibas]{rubner2000earth}
Yossi Rubner, Carlo Tomasi, and Leonidas~J Guibas.
\newblock The earth mover's distance as a metric for image retrieval.
\newblock \emph{International journal of computer vision}, 40\penalty0
  (2):\penalty0 99--121, 2000.

\bibitem[Schmidt and Jiang(2024)]{schmidt2023learning}
Dominik Schmidt and Minqi Jiang.
\newblock Learning to act without actions.
\newblock In \emph{International Conference on Learning Representations},
  volume 2024, pages 9379--9395, 2024.

\bibitem[Shen et~al.(2026)Shen, Huang, Kamanuru, Aliev, Rainton, Awelkair,
  Zeng, Li, Dong, Yuan, et~al.]{arxiv260710891}
Qijia Shen, Zhiqi Huang, Vamsidhar Kamanuru, Aznaur Aliev, Jay Rainton, Ahmed
  Awelkair, Zhichen Zeng, Jiajun Li, Shi Dong, Yueming Yuan, et~al.
\newblock Seta: Scaling environments for terminal agents.
\newblock \emph{arXiv preprint arXiv:2607.10891}, 2026.

\bibitem[Szot et~al.(2025)Szot, Mazoure, Attia, Timofeev, Agrawal, Hjelm, Gan,
  Kira, and Toshev]{szot2025multimodal}
Andrew Szot, Bogdan Mazoure, Omar Attia, Aleksei Timofeev, Harsh Agrawal, Devon
  Hjelm, Zhe Gan, Zsolt Kira, and Alexander Toshev.
\newblock From multimodal llms to generalist embodied agents: Methods and
  lessons.
\newblock In \emph{2025 IEEE/CVF Conference on Computer Vision and Pattern
  Recognition (CVPR)}, pages 10644--10655. IEEE, 2025.

\bibitem[Taesiri et~al.(2024)Taesiri, Feng, Bezemer, and
  Nguyen]{taesiri2023glitchbench}
Mohammad~Reza Taesiri, Tianjun Feng, Cor-Paul Bezemer, and Anh Nguyen.
\newblock Glitchbench: Can large multimodal models detect video game glitches?
\newblock In \emph{2024 IEEE/CVF Conference on Computer Vision and Pattern
  Recognition (CVPR)}, pages 22444--22455. IEEE, 2024.

\bibitem[Taesiri et~al.(2025)Taesiri, Ghildyal, Zadtootaghaj, Barman, and
  Bezemer]{taesiri2025videogameqa}
Mohammad~Reza Taesiri, Abhijay Ghildyal, Saman Zadtootaghaj, Nabajeet Barman,
  and Cor-Paul Bezemer.
\newblock Videogameqa-bench: Evaluating vision-language models for video game
  quality assurance.
\newblock \emph{Advances in Neural Information Processing Systems}, 38, 2025.

\bibitem[Tan et~al.(2024)Tan, Zhang, Xu, Xia, Ding, Li, Zhou, Yue, Jiang, Li,
  et~al.]{arxiv240303186}
Weihao Tan, Wentao Zhang, Xinrun Xu, Haochong Xia, Ziluo Ding, Boyu Li, Bohan
  Zhou, Junpeng Yue, Jiechuan Jiang, Yewen Li, et~al.
\newblock Cradle: Empowering foundation agents towards general computer
  control.
\newblock \emph{arXiv preprint arXiv:2403.03186}, 2024.

\bibitem[Tan et~al.(2025)Tan, Li, Fang, Yao, Yan, Luo, Ao, Li, Ren, Yi,
  et~al.]{arxiv251108892}
Weihao Tan, Xiangyang Li, Yunhao Fang, Heyuan Yao, Shi Yan, Hao Luo, Tenglong
  Ao, Huihui Li, Hongbin Ren, Bairen Yi, et~al.
\newblock Lumine: An open recipe for building generalist agents in 3d open
  worlds.
\newblock \emph{arXiv preprint arXiv:2511.08892}, 2025.

\bibitem[Thomas et~al.(2025)Thomas, Chan, Kang, Wu, Christianos, Greenlee,
  Toulis, and Purtorab]{arxiv250218356}
George Thomas, Alex~J Chan, Jikun Kang, Wenqi Wu, Filippos Christianos, Fraser
  Greenlee, Andy Toulis, and Marvin Purtorab.
\newblock Webgames: Challenging general-purpose web-browsing ai agents.
\newblock \emph{arXiv preprint arXiv:2502.18356}, 2025.

\bibitem[Tong et~al.(2026)Tong, Tang, Li, Mou, Zhang, Zhao, Wen, Song, Zhan,
  Lu, et~al.]{arxiv250513886}
Jingqi Tong, Jixin Tang, Hangcheng Li, Yurong Mou, Ming Zhang, Jun Zhao, Yanbo
  Wen, Fan Song, Jiahao Zhan, Yuyang Lu, et~al.
\newblock Game-rl: Synthesizing multimodal verifiable game data to boost vlms'
  general reasoning.
\newblock In \emph{International Conference on Learning Representations},
  volume 2026, pages 34689--34755, 2026.

\bibitem[Varghese et~al.(2026)Varghese, Vettoruzzo, Simoncini, Callejas,
  Derakhshani, Meding, Vanschoren, and Snoek]{arxiv261007127}
Dheeraj Varghese, Anna Vettoruzzo, Walter Simoncini, Michelle Lorena~Acevedo
  Callejas, Mohammad~Mahdi Derakhshani, Kristof Meding, Joaquin Vanschoren, and
  Cees~GM Snoek.
\newblock Playsuite: A large-scale benchmark for interactive visual
  intelligence.
\newblock \emph{arXiv preprint arXiv:2610.07127}, 2026.

\bibitem[Wang et~al.(2026{\natexlab{a}})Wang, Lu, Wang, Bai, Liu, Zhang, Wang,
  Hu, Xie, Bai, et~al.]{wang2026cuagym}
Bowen Wang, Dunjie Lu, Junli Wang, Tianyi Bai, Shixuan Liu, Zhipeng Zhang,
  Haiquan Wang, Hao Hu, Tianbao Xie, Shuai Bai, et~al.
\newblock Cua-gym: Scaling verifiable training environments and tasks for
  computer-use agents.
\newblock \emph{arXiv preprint arXiv:2605.25624}, 2026{\natexlab{a}}.

\bibitem[Wang et~al.(2023)Wang, Xie, Jiang, Mandlekar, Xiao, Zhu, Fan, and
  Anandkumar]{arxiv230516291}
Guanzhi Wang, Yuqi Xie, Yunfan Jiang, Ajay Mandlekar, Chaowei Xiao, Yuke Zhu,
  Linxi Fan, and Anima Anandkumar.
\newblock Voyager: An open-ended embodied agent with large language models.
\newblock \emph{arXiv preprint arXiv:2305.16291}, 2023.

\bibitem[Wang et~al.(2025{\natexlab{a}})Wang, Zou, Song, Feng, Fang, Lu, Liu,
  Luo, Liang, Huang, et~al.]{wang2025ui}
Haoming Wang, Haoyang Zou, Huatong Song, Jiazhan Feng, Junjie Fang, Junting Lu,
  Longxiang Liu, Qinyu Luo, Shihao Liang, Shijue Huang, et~al.
\newblock Ui-tars-2 technical report: Advancing gui agent with multi-turn
  reinforcement learning.
\newblock \emph{arXiv preprint arXiv:2509.02544}, 2025{\natexlab{a}}.

\bibitem[Wang et~al.(2019)Wang, Lehman, Clune, and Stanley]{arxiv190101753}
Rui Wang, Joel Lehman, Jeff Clune, and Kenneth~O Stanley.
\newblock Paired open-ended trailblazer (poet): Endlessly generating
  increasingly complex and diverse learning environments and their solutions.
\newblock \emph{arXiv preprint arXiv:1901.01753}, 2019.

\bibitem[Wang et~al.(2025{\natexlab{b}})Wang, Wang, Lu, Yang, Xie, Wang, Deng,
  Guo, Xu, Wu, et~al.]{wang2025opencua}
Xinyuan Wang, Bowen Wang, Dunjie Lu, Junlin Yang, Tianbao Xie, Junli Wang,
  Jiaqi Deng, Xiaole Guo, Yiheng Xu, Chen Wu, et~al.
\newblock Opencua: Open foundations for computer-use agents.
\newblock \emph{Advances in Neural Information Processing Systems},
  38:\penalty0 139756--139806, 2025{\natexlab{b}}.

\bibitem[Wang et~al.(2026{\natexlab{b}})Wang, Yin, Pu, Wang, Li, Ouyang, Sun,
  Li, Cheng, Yu, et~al.]{arxiv260925001}
Yiran Wang, Xingyilang Yin, Junfu Pu, Guangzhi Wang, Kaifeng Li, Mingyu Ouyang,
  Huiqiang Sun, Lingen Li, Cheng Cheng, Wangbo Yu, et~al.
\newblock Gamehorizon suite: Multi-horizon data and evaluation in gameplay.
\newblock \emph{arXiv preprint arXiv:2609.25001}, 2026{\natexlab{b}}.

\bibitem[Wang et~al.(2026{\natexlab{c}})Wang, Xu, Zheng, Zhang, Kogundi, Hans,
  and Ustun]{wang2026gameplayqa}
Yunzhe Wang, Runhui Xu, Kexin Zheng, Tianyi Zhang, Jayavibhav~Niranjan Kogundi,
  Soham Hans, and Volkan Ustun.
\newblock Gameplayqa: A benchmarking framework for decision-dense pov-synced
  multi-video understanding of 3d virtual agents.
\newblock In \emph{Proceedings of the 64th Annual Meeting of the Association
  for Computational Linguistics (Volume 1: Long Papers)}, pages 34333--34355,
  2026{\natexlab{c}}.

\bibitem[Wang et~al.(2025{\natexlab{c}})Wang, Li, Ye, Fang, Wang, Liu, Liang,
  Lu, Wu, Feng, et~al.]{arxiv251023691}
Zihao Wang, Xujing Li, Yining Ye, Junjie Fang, Haoming Wang, Longxiang Liu,
  Shihao Liang, Junting Lu, Zhiyong Wu, Jiazhan Feng, et~al.
\newblock Game-tars: Pretrained foundation models for scalable generalist
  multimodal game agents.
\newblock \emph{arXiv preprint arXiv:2510.23691}, 2025{\natexlab{c}}.

\bibitem[Weihs et~al.(2021)Weihs, Jain, Liu, Salvador, Lazebnik, Kembhavi, and
  Schwing]{weihs2020bridging}
Luca Weihs, Unnat Jain, Iou-Jen Liu, Jordi Salvador, Svetlana Lazebnik,
  Aniruddha Kembhavi, and Alex Schwing.
\newblock Bridging the imitation gap by adaptive insubordination.
\newblock \emph{Advances in Neural Information Processing Systems},
  34:\penalty0 19134--19146, 2021.

\bibitem[Wu et~al.(2026)Wu, Peng, Chen, Ruan, Zhuang, Yang, Zhang, Chen, Tseng,
  Yu, et~al.]{wu2026autowebworld}
Yifan Wu, Yiran Peng, Yiyu Chen, Jianhao Ruan, Zijie Zhuang, Cheng Yang, Jiayi
  Zhang, Man Chen, Yenchi Tseng, Zhaoyang Yu, et~al.
\newblock Autowebworld: Synthesizing infinite verifiable web environments via
  finite state machines.
\newblock \emph{arXiv preprint arXiv:2602.14296}, 2026.

\bibitem[Wu et~al.(2024)Wu, Tang, Mitchell, and Li]{arxiv231001557}
Yue Wu, Xuan Tang, Tom Mitchell, and Yuanzhi Li.
\newblock Smartplay: A benchmark for llms as intelligent agents.
\newblock In \emph{International Conference on Learning Representations},
  volume 2024, pages 1543--1561, 2024.

\bibitem[Xie et~al.(2026)Xie, Ma, Lan, Yuille, Xiao, and Wei]{arxiv250608011}
Yunfei Xie, Yinsong Ma, Shiyi Lan, Alan Yuille, Junfei Xiao, and Chen Wei.
\newblock Play to generalize: Learning to reason through game play.
\newblock In \emph{International Conference on Learning Representations},
  volume 2026, pages 17213--17243, 2026.

\bibitem[Xu et~al.(2024)Xu, Wang, Wang, Lu, Xie, Saha, Sahoo, Yu, and
  Xiong]{xu2024aguvis}
Yiheng Xu, Zekun Wang, Junli Wang, Dunjie Lu, Tianbao Xie, Amrita Saha, Doyen
  Sahoo, Tao Yu, and Caiming Xiong.
\newblock Aguvis: Unified pure vision agents for autonomous gui interaction.
\newblock \emph{arXiv preprint arXiv:2412.04454}, 2024.

\bibitem[Xu et~al.(2026)Xu, Xu, Yi, Yuan, Guang, Long, Chen, Wu, Yu, and
  Wang]{arxiv250602387}
Zelai Xu, Zhexuan Xu, Xiangmin Yi, Huining Yuan, Mo~Guang, Kaiwen Long, Xinlei
  Chen, Yi~Wu, Chao Yu, and Yu~Wang.
\newblock Vs-bench: Evaluating vlms for strategic abilities in multi-agent
  environments.
\newblock In \emph{Proceedings of the IEEE/CVF Conference on Computer Vision
  and Pattern Recognition}, pages 21561--21572, 2026.

\bibitem[Xu et~al.(2025)Xu, Wang, Li, Yu, Wang, Fu, and Yang]{arxiv250313356}
Zhongwen Xu, Xianliang Wang, Siyi Li, Tao Yu, Liang Wang, Qiang Fu, and Wei
  Yang.
\newblock Agents play thousands of 3d video games.
\newblock \emph{arXiv preprint arXiv:2503.13356}, 2025.

\bibitem[Yang et~al.(2025{\natexlab{a}})Yang, Lieret, Jimenez, Wettig,
  Khandpur, Zhang, Hui, Press, Schmidt, and Yang]{yang2025swesmith}
John Yang, Kilian Lieret, Carlos Jimenez, Alexander Wettig, Kabir Khandpur,
  Yanzhe Zhang, Binyuan Hui, Ofir Press, Ludwig Schmidt, and Diyi Yang.
\newblock Swe-smith: Scaling data for software engineering agents.
\newblock \emph{Advances in Neural Information Processing Systems}, 38,
  2025{\natexlab{a}}.

\bibitem[Yang et~al.(2025{\natexlab{b}})Yang, Chen, Zhang, Zhao, Qian, Wang,
  Wang, Koripella, Movahedi, Li, et~al.]{yang2025embodiedbench}
Rui Yang, Hanyang Chen, Junyu Zhang, Mark Zhao, Cheng Qian, Kangrui Wang,
  Qineng Wang, Teja~Venkat Koripella, Marziyeh Movahedi, Manling Li, et~al.
\newblock Embodiedbench: Comprehensive benchmarking multi-modal large language
  models for vision-driven embodied agents.
\newblock \emph{arXiv preprint arXiv:2502.09560}, 2025{\natexlab{b}}.

\bibitem[Yang et~al.(2026{\natexlab{a}})Yang, Wu, Chen, Bai, Yao, Cheng, Peng,
  Zhang, Zhang, and Gao]{yang2026openwebrl}
Rui Yang, Qianhui Wu, Yuxi Chen, Hao Bai, Wenlin Yao, Hao Cheng, Baolin Peng,
  Huan Zhang, Tong Zhang, and Jianfeng Gao.
\newblock Openwebrl: Demystifying online multi-turn reinforcement learning for
  visual web agents.
\newblock \emph{arXiv preprint arXiv:2606.02031}, 2026{\natexlab{a}}.

\bibitem[Yang et~al.(2026{\natexlab{b}})Yang, Wu, Wang, Chen, Yang, Cheng, Yao,
  Peng, Zhang, Gao, et~al.]{yang2026gui}
Rui Yang, Qianhui Wu, Zhaoyang Wang, Hanyang Chen, Ke~Yang, Hao Cheng, Huaxiu
  Yao, Baolin Peng, Huan Zhang, Jianfeng Gao, et~al.
\newblock Gui-libra: Training native gui agents to reason and act with
  action-aware supervision and partially verifiable rl.
\newblock \emph{arXiv preprint arXiv:2602.22190}, 2026{\natexlab{b}}.

\bibitem[Yang et~al.(2024)Yang, Li, Yan, Zhang, Huang, and Liu]{yang2024react}
Zonghan Yang, Peng Li, Ming Yan, Ji~Zhang, Fei Huang, and Yang Liu.
\newblock React meets actre: When language agents enjoy training data autonomy.
\newblock \emph{arXiv preprint arXiv:2403.14589}, 2024.

\bibitem[Yao et~al.(2022)Yao, Zhao, Yu, Du, Shafran, Narasimhan, and
  Cao]{yao2022react}
Shunyu Yao, Jeffrey Zhao, Dian Yu, Nan Du, Izhak Shafran, Karthik Narasimhan,
  and Yuan Cao.
\newblock React: Synergizing reasoning and acting in language models.
\newblock \emph{arXiv preprint arXiv:2210.03629}, 2022.

\bibitem[Ying et~al.(2026)Ying, Truong, Sharma, Zhao, Cloos, Allen, Griffiths,
  Collins, Hern{\'a}ndez-Orallo, Isola, et~al.]{arxiv260217594}
Lance Ying, Ryan Truong, Prafull Sharma, Kaiya~Ivy Zhao, Nathan Cloos, Kelsey~R
  Allen, Thomas~L Griffiths, Katherine~M Collins, Jos{\'e}
  Hern{\'a}ndez-Orallo, Phillip Isola, et~al.
\newblock Ai gamestore: Scalable, open-ended evaluation of machine general
  intelligence with human games.
\newblock \emph{arXiv preprint arXiv:2602.17594}, 2026.

\bibitem[Yue et~al.(2025)Yue, Green, Hunt, Salia, Shi, and Hunt]{yue2025pixels}
Yuguang Yue, Chris Green, Samuel Hunt, Irakli Salia, Wenzhe Shi, and Jonathan~J
  Hunt.
\newblock Pixels to play: A foundation model for 3d gameplay.
\newblock In \emph{2025 IEEE Conference on Games (CoG)}, pages 1--4. IEEE,
  2025.

\bibitem[Yue et~al.(2026)Yue, Salia, Hunt, Green, Shi, and
  Hunt]{yue2026scaling}
Yuguang Yue, Irakli Salia, Samuel Hunt, Chris Green, Wenzhe Shi, and Jonathan~J
  Hunt.
\newblock Scaling behavior cloning improves causal reasoning: An open model for
  real-time video game playing.
\newblock \emph{arXiv preprint arXiv:2601.04575}, 2026.

\bibitem[Zala et~al.(2024)Zala, Cho, Lin, Yoon, and Bansal]{zala2024envgen}
Abhay Zala, Jaemin Cho, Han Lin, Jaehong Yoon, and Mohit Bansal.
\newblock {EnvGen}: Generating and adapting environments via {LLMs} for
  training embodied agents.
\newblock In \emph{Conference on Language Modeling}, 2024.

\bibitem[Zawalski et~al.(2024)Zawalski, Chen, Pertsch, Mees, Finn, and
  Levine]{zawalski2024robotic}
Micha{\l} Zawalski, William Chen, Karl Pertsch, Oier Mees, Chelsea Finn, and
  Sergey Levine.
\newblock Robotic control via embodied chain-of-thought reasoning.
\newblock \emph{arXiv preprint arXiv:2407.08693}, 2024.

\bibitem[Zhang et~al.(2025)Zhang, Griffiths, Narasimhan, and
  Press]{arxiv250518134}
Alex~L Zhang, Thomas~L Griffiths, Karthik~R Narasimhan, and Ofir Press.
\newblock Videogamebench: Can vision-language models complete popular video
  games?
\newblock \emph{arXiv preprint arXiv:2505.18134}, 2025.

\bibitem[Zhang et~al.(2026{\natexlab{a}})Zhang, Chen, Yang, Su, Wu, Chen, Luo,
  Hou, Tan, Zheng, et~al.]{zhang2026gamescaling}
Kuan Zhang, Yukun Chen, Zhihao Yang, Yue Su, Xiangnan Wu, Zirong Chen, Run Luo,
  Jinkun Hou, Tao Tan, Yinhe Zheng, et~al.
\newblock Scaling in games: Continued pre-training for embodied agents in
  diverse virtual worlds.
\newblock \emph{arXiv preprint arXiv:2610.28698}, 2026{\natexlab{a}}.

\bibitem[Zhang et~al.(2026{\natexlab{b}})Zhang, Liu, Zhao, Hou, Zhang, Xie,
  Liu, and Li]{arxiv260306656}
Kuan Zhang, Dongchen Liu, Qiyue Zhao, Jinkun Hou, Xinran Zhang, Qinlei Xie,
  Miao Liu, and Yiming Li.
\newblock Gameverse: Can vision-language models learn from video-based
  reflection?
\newblock \emph{arXiv preprint arXiv:2603.06656}, 2026{\natexlab{b}}.

\bibitem[Zhang et~al.(2020)Zhang, Walshe, Liu, Guan, Muller, Whritner, Zhang,
  Hayhoe, and Ballard]{zhang2019atari}
Ruohan Zhang, Calen Walshe, Zhuode Liu, Lin Guan, Karl Muller, Jake Whritner,
  Luxin Zhang, Mary Hayhoe, and Dana Ballard.
\newblock Atari-head: Atari human eye-tracking and demonstration dataset.
\newblock In \emph{Proceedings of the AAAI conference on artificial
  intelligence}, volume~34, pages 6811--6820, 2020.

\end{thebibliography}
